\documentclass[letterpaper]{article} 
\usepackage{aaai25}  

\usepackage{times}  
\usepackage{helvet}  
\usepackage{courier}  
\usepackage[hyphens]{url}  
\usepackage{graphicx} 
\usepackage{relsize}
\usepackage{natbib}  
\usepackage{caption} 
\usepackage{algorithm}
\usepackage{algorithmic}

\usepackage{newfloat}
\usepackage{listings}
\DeclareCaptionStyle{ruled}{labelfont=normalfont,labelsep=colon,strut=off} 
\floatstyle{ruled}
\newfloat{listing}{tb}{lst}{}
\floatname{listing}{Listing}
\title{On Local Overfitting and Forgetting in Deep Neural Networks}
\author{
    Uri Stern, Tomer Yaacoby and Daphna Weinshall
}
\affiliations{

    School of Computer Science and Engineering, The Hebrew University of Jerusalem, Jerusalem 91904, Israel\\
    ustern@gmail.com, tomer.yaacoby@mail.huji.ac.il, daphna@mail.huji.ac.il\\
}

\usepackage{xcolor}         
\usepackage[font={small}]{caption}
\usepackage[font={scriptsize}]{subcaption}
\usepackage{paralist}
\usepackage{enumitem}
\usepackage{sidecap}
\usepackage{amssymb}
\usepackage{amsmath}
\usepackage{mathtools}
\usepackage{amsthm}
\usepackage{amsfonts}
\usepackage{bm}
\usepackage{booktabs}

\newcommand{\bx}{{\mathbf x}}

\newcommand{\by}{\bm{y}}
\newcommand{\bw}{\bm{w}}

\newcommand{\bW}{\bm{w}}

\newcommand{\R}{\mathbb{R}}

\newcommand{\Sc}{\mathcal{S}}

\newcommand{\Mc}{\mathcal{M}}
\newcommand{\hW}{\bm{W}}

\newcommand{\myparagraph}[1]{\smallskip\noindent\textbf{#1}\;}

\newtheorem{claim}{Claim}

\newtheorem{defn}{Definition}

\newtheorem*{result*}{Theorem}

\newcommand{\app}{{App.~}}

\begin{document}

\maketitle

\begin{abstract}

The infrequent occurrence of overfitting in deep neural networks is perplexing: contrary to theoretical expectations, increasing model size often enhances performance in practice. But what if overfitting does occur, though restricted to specific sub-regions of the data space? In this work, we propose a novel score that captures the forgetting rate of deep models on validation data. We posit that this score quantifies \emph{local overfitting}: a decline in performance confined to certain regions of the data space. We then show empirically that \emph{local overfitting} occurs regardless of the presence of traditional overfitting. Using the framework of deep over-parametrized linear models, we offer a certain theoretical characterization of forgotten knowledge, and show that it correlates with knowledge forgotten by real deep models. Finally, we devise a new ensemble method that aims to recover forgotten knowledge, relying solely on the training history of a single network. When combined with self-distillation, this method will enhance the performance of a trained model without adding inference costs. Extensive empirical evaluations demonstrate the efficacy of our method across multiple datasets, contemporary neural network architectures, and training protocols.

\end{abstract}

\section{Introduction}

Overfitting a training set is considered a fundamental challenge in machine learning. Theoretical analyses predict that as a model gains additional degrees of freedom, its capacity to fit a given training dataset increases. Consequently, there is a point where the model becomes too specialized for a particular training set, leading to an increase in its generalization error. In deep learning, one would expect to see \emph{increased generalization error} as the number of parameters and/or training epochs increases. Surprisingly, even vast deep neural networks with billions of parameters seldom adhere to this expectation, and overfitting as a function of epochs is almost never observed \citep{liu2022convnet}. Typically, a significant increase in the number of parameters still results in enhanced performance, or occasionally in peculiar phenomena like the double descent in test error \citep{annavarapu2021deep}, see Section~\ref{sec:overfitanddoubledescent}. Clearly, there exists a gap between our classical understanding of overfitting and the empirical results observed when training modern neural networks.

\begin{figure}[tb]
    \center
    \includegraphics[width=.38\linewidth]{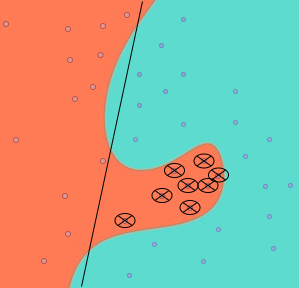} \hspace{.5cm}
    \includegraphics[width=.38\linewidth]{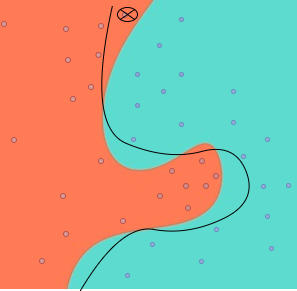} 
    \caption{Local overfitting and forgetting in a binary problem, where blue and orange denote the different classes, and circles mark the validation set. The initial (left) and final (right) separators of a hypothetical learning method are shown, where $\otimes$ marks prediction errors. Clearly the final classifier has a smaller generalization error, but now one point at the top is 'forgotten'.}
\vspace{-.5cm}
\label{fig:illustration}
\end{figure}

To bridge this gap, we present a fresh perspective on overfitting. Instead of solely assessing it through a decline in \emph{validation accuracy}, we propose to monitor what we term \emph{the model's forget fraction}. This metric quantifies the portion of test data (or validation set) that the model initially classifies correctly but misclassifies as training proceeds (see illustration in Fig.~\ref{fig:illustration}). Throughout this paper we term the decline in test accuracy as "forgetting", to emphasize that the model's ability to correctly classify portions of the data is reduced. In Section~\ref{sec:overfitanddoubledescent}, we investigate various benchmark datasets, observing this phenomenon even in the absence of overfitting as conventionally defined, i.e., when test accuracy increases throughout. 
Notably, this occurs in competitive networks despite the implementation of modern techniques to mitigate overfitting, such as data augmentation and dropout. Our empirical investigation also reveals that forgetting of patterns occurs alongside the learning of new patterns in the training set, explaining why the traditional definition of overfitting fails to capture this phenomenon.

Formal investigation of the phenomenon of forgotten knowledge is challenging, particularly in the context of deep neural networks which are not easily amenable to formal analysis. Instead, in Section~\ref{sec:ForgottenKnowledge} we adopt the framework of over-parameterized deep linear networks. This framework involves non-linear optimization and has previously offered valuable insights into the learning processes of practical deep networks \citep{fukumizu1998effect,DBLP:journals/corr/SaxeMG13,arora2018optimization,DBLP:conf/iclr/AroraCGH19,DBLP:conf/iclr/HuXP20}. Within such models, employing gradient descent for learning reveals a straightforward and elegant characterization of the model's evolution \citep{hacohen2022principal}.

We expand upon this analysis, deriving an analytical description of the data points forgotten at each gradient descent step. As this analysis pertains specifically to deep linear models, it's crucial to correlate its findings with the forgotten knowledge in competitive neural networks. Intriguingly, when comparing these findings with the same image datasets utilized in our experiments, we observe significant overlap between the sets. This implies that the proposed theoretical characterization might offer partial insight into the phenomenon of forgotten knowledge and the underlying causes of local overfitting.

Based on the empirical observations reported in Section~\ref{sec:overfitanddoubledescent}, we propose in Section~\ref{sec:method} a method that can effectively reduce the forgetting of test data, and thus improve the final accuracy and reduce overfitting. More specifically, we propose a new prediction method that combines knowledge gained in different stages of training. The method delivers a weighted average of the class probability output vector between the final model and a set of checkpoints of the model from mid-training, where the checkpoints and their weights are chosen in an iterative manner using a validation dataset and our forget metric. 
The purpose is two-fold: First, an improvement upon the original model by our method will serve as another strong indication that models indeed forget useful knowledge in the late stages of training. Second, to provide a proof-of-concept that this lost knowledge can be recovered, even with methods as simple as ours. 

In Section~\ref{sec:empirical} we describe the empirical validation of our method in a series of experiments over image classification datasets with and without label noise, using various network architectures, including in particular modern networks over Imagenet. The results indicate that our method is universally useful and generally improves upon the original model, thus fulfilling its two mentioned goals. When compared with alternative methods that use the network's training history, our method shows comparable or improved performance, while being more general and easy to use (both in implementation and hyper-parameter tuning). Unlike some methods, it does not depend on additional training choices that require much more time and effort to tune the new hyper-parameters. 

\myparagraph{Our main contributions.}  \begin{inparaenum}[(i)] \item A novel perspective on overfitting, capturing the notion of \emph{local overfitting}. \item Empirical evidence that overfitting occurs \textbf{locally} even without a decrease in overall generalization. \item Analysis of the relation between forgetting and PCA. \item A simple and effective method to reduce overfitting, and its empirical validation.\end{inparaenum}

\section{Related Work}

\myparagraph{Study of forgetting in prior work.} Most existing studies examine the forgetting of training data, where certain training points are initially memorized but later forgotten. This typically occurs when the network cannot fully memorize the training set. In contrast, our work focuses on the \emph{forgetting of validation points}, which arises when the network successfully memorizes the entire training set. Building on \citet{arpit2017closer}, who show that networks first learn "simple" patterns before transitioning to memorizing noisy data, we analyze the later stages of learning, particularly in the context of the double descent phenomenon. Another related but distinct phenomenon is "catastrophic forgetting" \citep{mccloskey1989catastrophic}, which occurs in \emph{continual learning} settings where the training data evolves over time—unlike the static training scenario considered here.
 
\myparagraph{Ensemble learning.} Ensemble learning has been studied extensively \citep[see][]{polikar2012ensemble, ganaie2022ensemble, yang2023survey}. Our work belongs to a line of works called 
"implicit ensemble learning", in which only a single network is learned in a way that "mimics" ensemble learning \citep{srivastava2014dropout}. Utilizing checkpoints from the training history as a 'cost-effective' ensemble has also been considered. This was achieved by either considering the last epochs and averaging their probability outputs \citep{xie2013horizontal}, or by employing exponential moving average (EMA) on all the weights throughout training \citep{polyak1992acceleration}. The latter method does not always succeed in reducing overfitting, as discussed in \citep{izmailov2018averaging}.

Several methods \citep{izmailov2018averaging, garipov2018loss, huang2017snapshot} modify the training protocol to converge to multiple local minima, which are then combined into an ensemble classifier. While these approaches show promise \citep{noppitak2022dropcyclic}, they add complexity to training and may even hurt performance \citep{guo2023stochastic}. Our comparisons (see Table~\ref{table:specialmethods}) demonstrate that our simpler method either matches or outperforms these techniques in all studied cases.

Ensemble methods can impose significant computational demands during inference, especially for large ensembles. Self-distillation \citep{allenzhu2023} addresses this challenge by training a single student model to replicate the ensemble’s predictions, effectively eliminating the increased computational costs. This approach typically maintains the ensemble’s performance and, in high-noise scenarios, may outperform the ensemble itself \cite{jeong2024understanding,stern2024united}.

\myparagraph{Studies of overfitting and double descent.} 
Double descent with respect to model size has been studied empirically in  \citep{belkin2019reconciling, nakkiran2021deep}, while epoch-wise double descent (which is the phenomenon analyzed here) was studied in \citep{stephenson2021and, heckel2020early}. These studies analyzed when and how epoch-wise double descent occurs, specifically in data with label noise, and explored ways to avoid it (sometimes at the cost of reduced generalization). Essentially, our research identifies a similar phenomenon in data without label noise. It is complementary to the study of "benign overfitting", e.g., the fact that models can achieve perfect fit to the train data while still obtaining good performance over the test data.

\section{Overfitting Revisited}
\label{sec:overfitanddoubledescent}

\begin{figure*}[htbp]
    \centering
    \begin{subfigure}[b]{0.24\linewidth}
        \includegraphics[width=\linewidth]{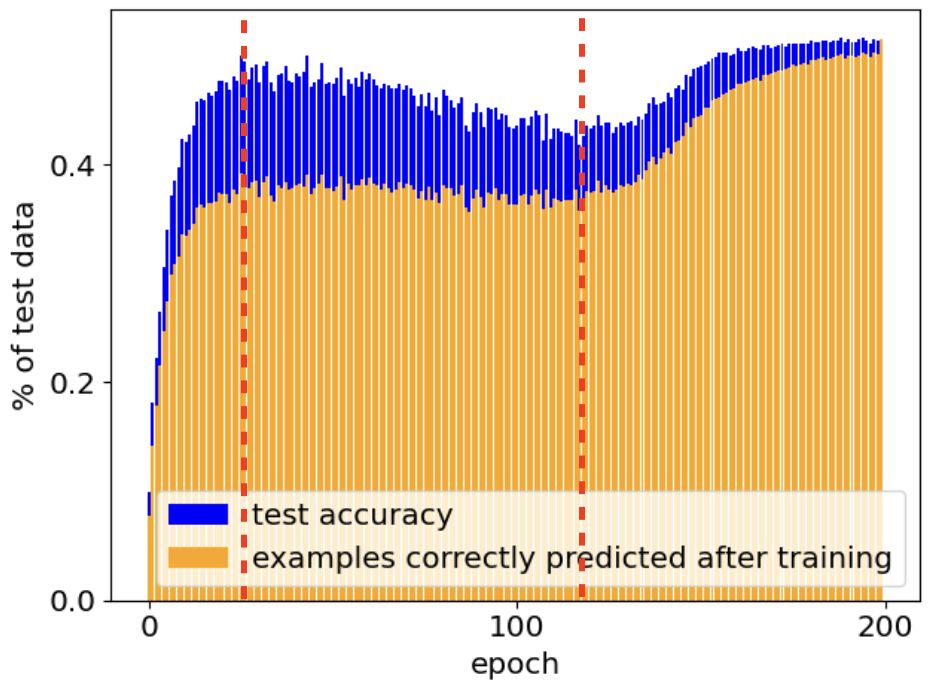}
    \vspace{-.6cm}
        \caption{Cifar100, 20\% sym noise}
        \label{subfig:DDc100sym20}
    \end{subfigure}
    \begin{subfigure}[b]{0.24\linewidth}
        \includegraphics[width=\linewidth]{figures/newtestbarTimgsym20r18.png}
    \vspace{-.6cm}
        \caption{TinyImagenet, 20\% sym noise}
        \label{subfig:DDTimgsym20}
    \end{subfigure}
    \begin{subfigure}[b]{0.24\linewidth}
        \includegraphics[width=\linewidth]{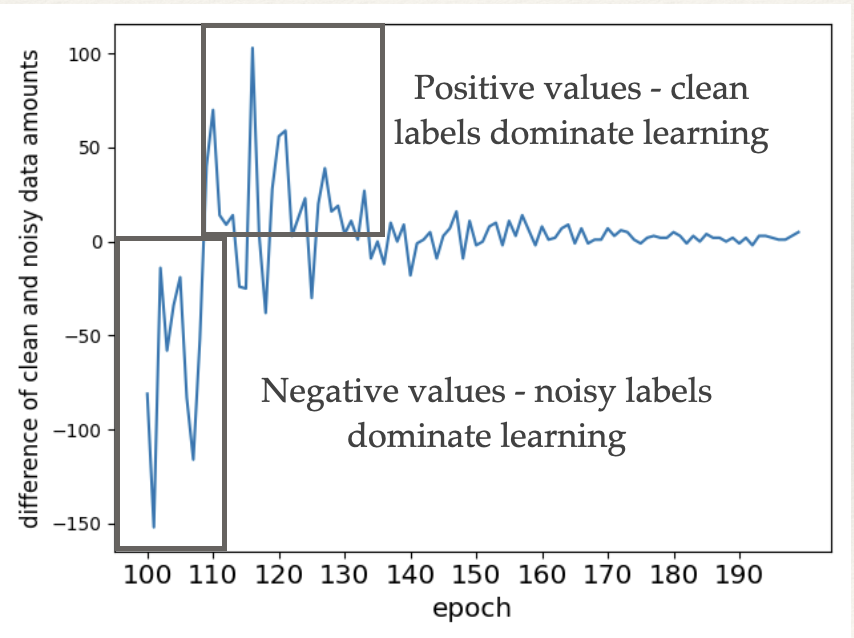}
    \vspace{-.6cm}
        \caption{Cifar100, 20\% sym noise}
        \label{subfig:c100sym20testacc}
    \end{subfigure}
    \begin{subfigure}[b]{0.24\linewidth}
        \includegraphics[width=\linewidth]{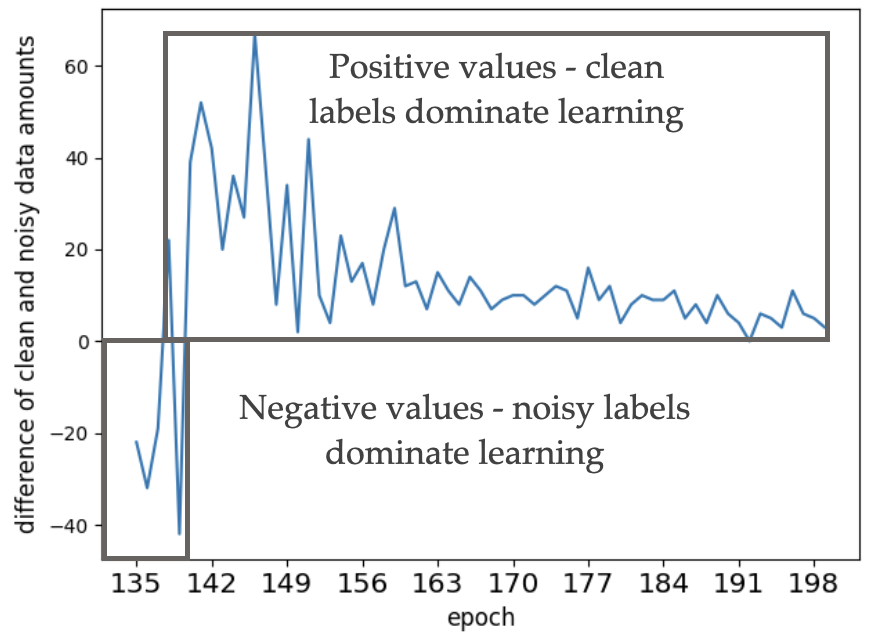}
    \vspace{-.6cm}
        \caption{TinyImagenet, 20\% sym noise}
        \label{subfig:logitsheatmapc100}
    \end{subfigure}
\vspace{-0.2cm}
     \caption[heatmap]{(a)-(b): Blue denotes test accuracy. Among those correctly recognized in each epoch $e$, orange denotes the fraction that remains correctly recognized at the end. The test accuracy (the blue curve) shows a clear double ascent of accuracy, which is much less pronounced in the orange curve. During the decrease in test accuracy - the range of epochs between the first and second dashed red vertical lines - the large gap between the blue and orange plots indicates the fraction of test data that has been correctly learned in the first ascent and then forgotten, without ever being re-learned in the later recovery period of the second ascent. (c)-(d): The difference between the number of clean and noisy datapoints at each epoch during the second ascent of test accuracy (the epochs after the second dashed red vertical line), counting datapoints with large loss only. Positive (negative) value indicates that clean (noisy) datapoints are more dominant in the corresponding epoch.
     }   
     \label{fig:doubledescent}
\vspace{-0.2cm}
\end{figure*}

The textbook definition of overfitting entails the co-occurrence of increasing train accuracy and decreasing generalization. Let $acc(e,S)$ denote the accuracy over set $S$ in epoch $e$ - some epoch in mid-training, $E$ the total number of epochs, and $T$ the test\footnote{Below, 'test set' and 'validation set' are used interchangeably.} dataset. Using test accuracy to approximate generalization, this implies that overfitting occurs at epoch $e$ when $acc(e,T) \geq acc(E,T)$.  

We begin by investigating the hypothesis that portions of the test data $T$ may be forgotten by the network during training. 
When we examine  the 'epoch-wise double descent', which frequently occurs during training on datasets with significant label noise, we indeed observe that a notable forgetting of the test data coincides with the memorization of noisy labels. Here, forgetting serves as an objective indicator of overfitting. When we further examine the training of modern networks on standard datasets (devoid of label noise), where overfitting (as traditionally defined) is absent, we discover a similar phenomenon (though of weaker magnitude): \emph{the networks still appear to forget certain sub-regions of the test population}. This observation, we assert, signifies a significant and more subtle form of overfitting in deep learning.

\myparagraph{Local overfitting.} Let $M_{e}$ denote the subset of the test data \emph{mislabeled} by the network at some epoch $e$. We define below two scores $L_e$ and $F_e$:
\begin{equation}
\label{eq:forget}
    F_e = \frac{acc(e,M_{E})\cdot|M_{E}|}{|T|}, ~
    L_e = \frac{acc(E,M_{e})\cdot|M_{e}|}{|T|}
\end{equation}
The \emph{forget fraction} $F_e$ represents the fraction of test points correctly classified at epoch e but misclassified by the final model. $L_e$ represents the fraction of test points misclassified at epoch $e$ but correctly classified by the final model. The relationship $acc(E, T) = acc(e, T)+L_e-F_e$ follows\footnote{$acc(E,T) - L_{e} = acc(e,T) - F_{e}$ is the fraction of test points correctly classified in both $e$ and $E$.}. In line with the classical definition of overfitting, if $L_{e} < F_{e}$, overfitting occurs since $acc(E, T) < acc(e, T)$.

But what if $L_e \geq F_e~\forall e$? By its classical definition \emph{overfitting does not occur} since the test accuracy increases continuously. Nevertheless, there may still be local overfitting as defined above, since $F_e>0$ indicates that data has been forgotten even if $L_e \geq F_e$.

\myparagraph{Reflections on the epoch-wise double descent.} Epoch-wise double descent (see Fig.~\ref{fig:doubledescent}) is an empirical observation \citep{belkin2019reconciling}, which shows that neural networks can improve their performance even after overfitting, thus causing \emph{double descent in test error} during training (note that we show the corresponding \emph{double-ascent in test accuracy}). This phenomenon is characteristic of learning from data with label noise, and is strongly related to overfitting since the dip in test accuracy co-occurs with the memorization of noisy labels.

We examine the behavior of score $F_e$ in this context and make a novel observation: when we focus on the fraction of data correctly classified by the network during the second rise in test accuracy, we observe that the data newly memorized during these epochs often differs from the data forgotten during the overfitting phase (the dip in accuracy). In fact, most of this data has been previously misclassified (see Figs.~\ref{subfig:DDc100sym20}-\ref{subfig:DDTimgsym20}). Figs.\ref{subfig:c100sym20testacc}-\ref{subfig:logitsheatmapc100} further illustrate that during the later stages of training on data with label noise, the majority of the data being memorized is, in fact, data with clean labels, which explains the second increase in test accuracy. It thus appears that epoch-wise double descent is caused by the \emph{simultaneous learning} of general (but hard to learn) patterns from clean data, and irrelevant features of noisy data.

\myparagraph{Forgetting in the absence of label noise} When training deep networks on visual benchmark datasets without added label noise, double descent  rarely occurs, if ever. In contrast, we observe that local overfitting, as captured by our new score $F_e$, commonly occurs.

\begin{figure}[htb]
    \centering
    \begin{subfigure}[b]{0.325\linewidth}
\vspace{-.1cm}
        \includegraphics[width=\linewidth]{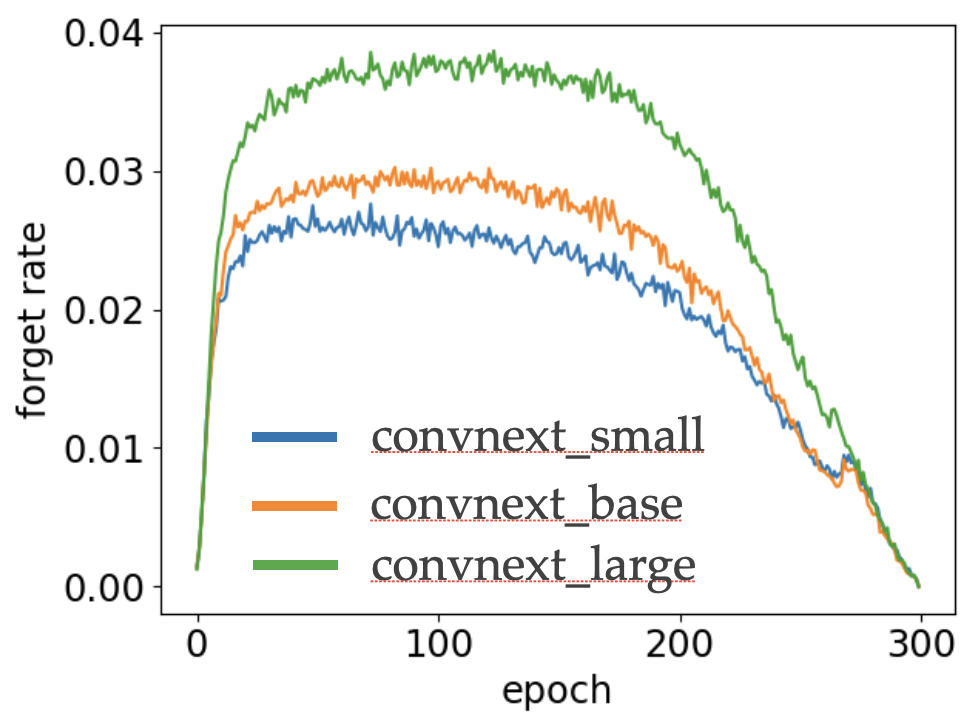}
    \vspace{-.5cm}
        \caption{ImageNet}
        \label{subfig:modelsizeforget}
    \end{subfigure}
    \begin{subfigure}[b]{0.325\linewidth}
\vspace{-.1cm}
        \includegraphics[width=\linewidth]{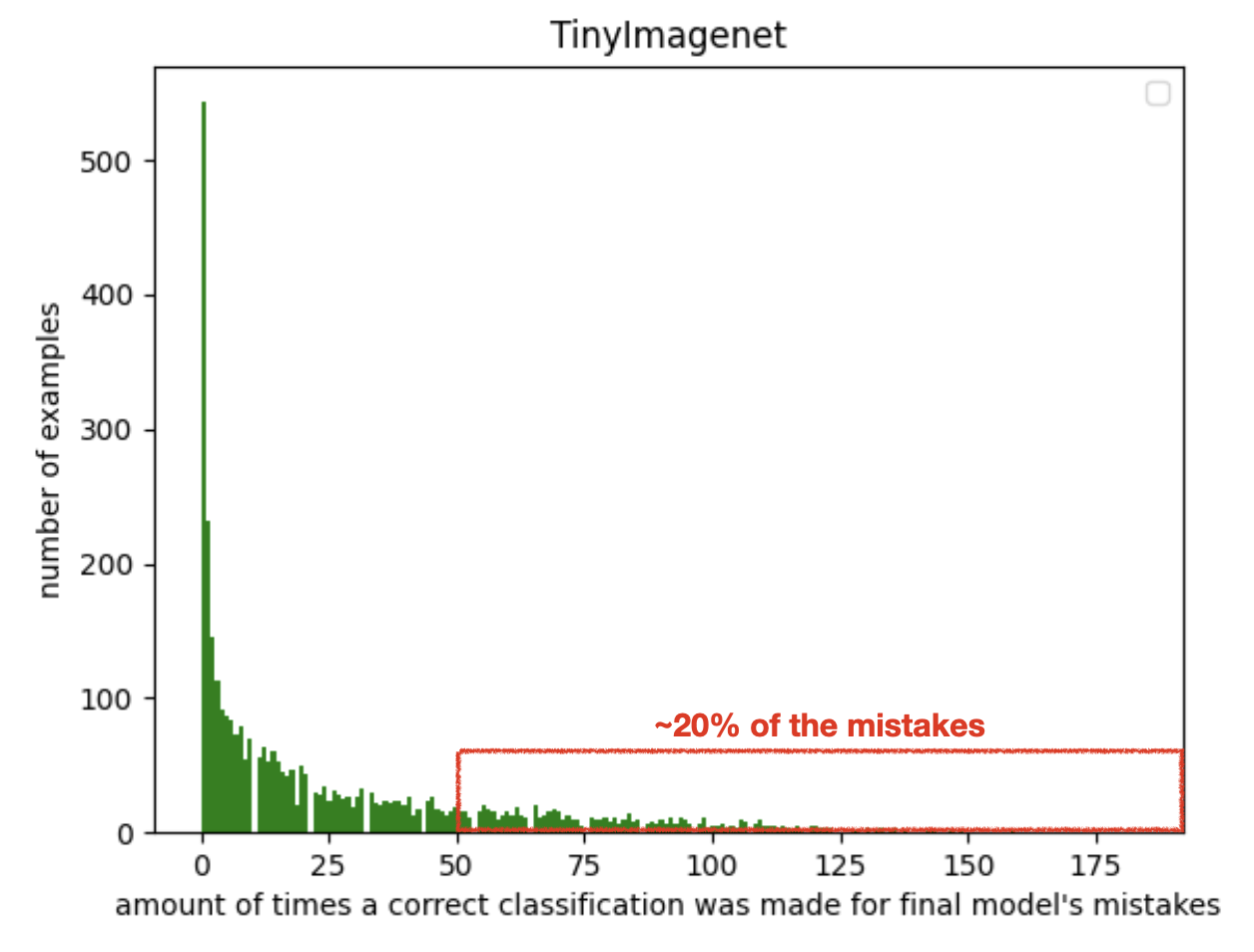}
    \vspace{-.5cm}
        \caption{TinyImagenet, resnet 18}
        \label{subfig:Timgsumcorrect}
    \end{subfigure}
    \begin{subfigure}[b]{0.325\linewidth}
\vspace{-.1cm}
        \includegraphics[width=\linewidth]{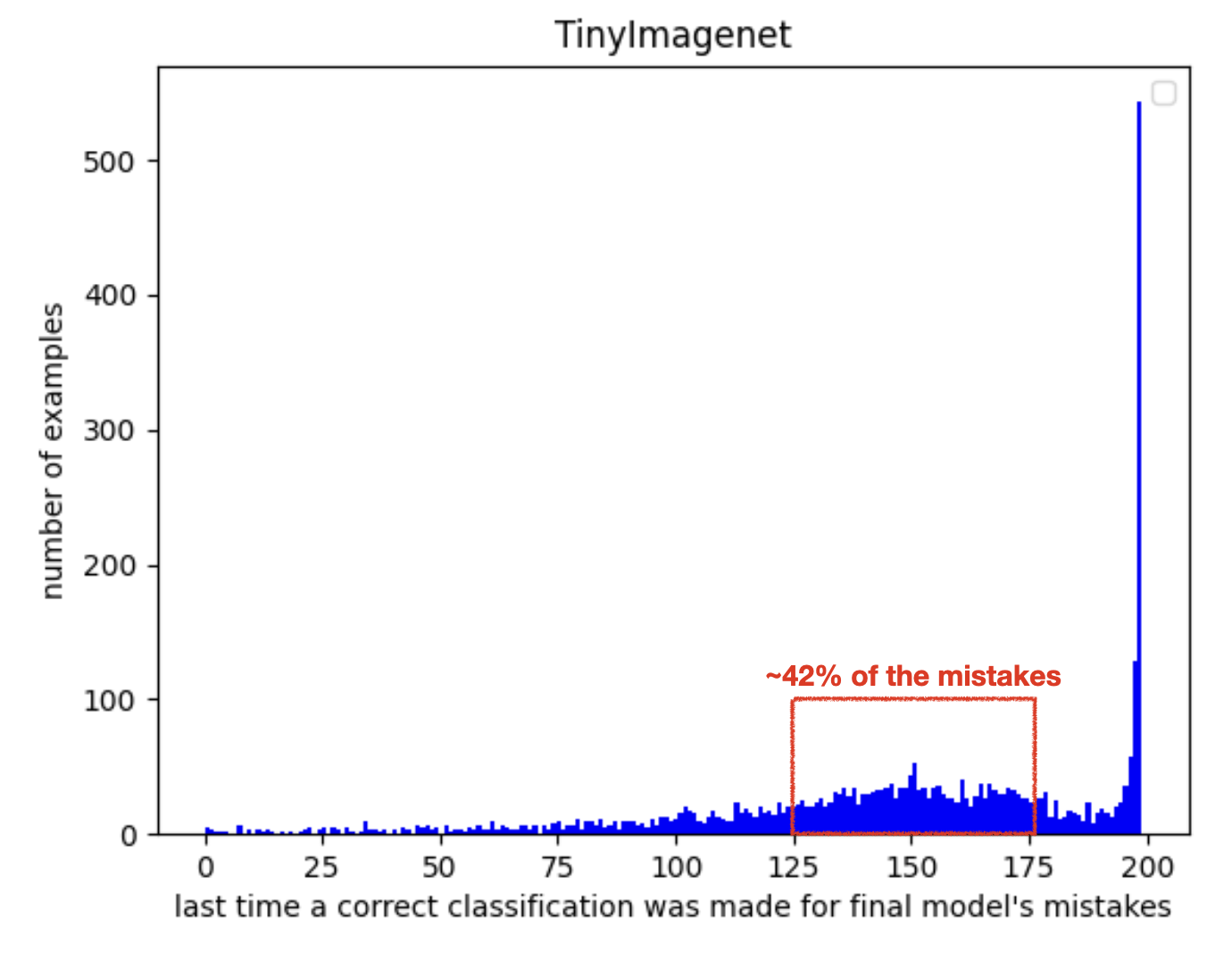}
    \vspace{-.5cm}
        \caption{TinyImagenet, resnet 18}
        \label{subfig:forgettimetimg}
    \end{subfigure}
     \caption[forget]{(a) The $F_e$ score (\ref{eq:forget}) of ConvNeXt trained on Imagenet, 3 network sizes: small $\rightarrow$ blue, base $\rightarrow$ orange and large $\rightarrow$ green. Accuracy remained consistent across all network sizes, while it is evident that $F_e$ increases with the network size. 
     (b-c) Within the set of wrongly classified test points after training, we show (b) the fraction that was correctly predicted (y-axis) for x epochs (x-axis), and (c) the last epoch in which an example was classified correctly.} 
\end{figure}

To show this, we trained various neural networks (ConvNets: Resnet, ConvNeXt; Visual transformers: ViT, MaxViT) on various datasets (CIFAR-100, TinyImagenet, Imagenet) using a variety of optimizers (SGD, AdamW) and learning rate schedulers (cosine annealing, steplr). In Fig.~\ref{subfig:modelsizeforget} we report the results, showing that all networks forget some portion of the data during training as in the label noise scenario, even if the test accuracy never decreases. Figs.~\ref{subfig:Timgsumcorrect}-\ref{subfig:forgettimetimg} demonstrate that this effect is not simply due to random fluctuations: many test examples that are incorrectly classified post training have been correctly classified during much of the training. These results are connected to overfitting in Fig.~\ref{subfig:modelsizeforget}: when investigating larger models and/or relatively small amounts of train data, which are scenarios that are expected to increase overfitting based on theoretical considerations, we see larger \emph{forget fraction} $F_e$ (see Figs.~\ref{fig:moreforgetrate}-\ref{fig:forgetrate} in \app\ref{app:moreforget}.

\myparagraph{In summary,} we see that neural networks can, and often will, "forget" significant portions of the test population as their training proceeds. In a sense, the networks \emph{are} overfitting, but this only occurs at some limited sub-regions of the world. The reason this failing is not captured by the classical definition of overfitting is that the networks continue to learn new general patterns simultaneously. In Section~\ref{sec:method} we discuss \emph{how we can harness this observation to improve the network's performance}.

\section{Forgotten Knowledge: Theory \& Exps} 
\label{sec:ForgottenKnowledge}

To gain insight into the nature of knowledge forgotten while training a deep model with Gradient Descent (GD), we analyze over-parameterized deep linear networks trained by GD. These models are constructed through the concatenation of linear operators in a multi-class classification scenario: $\by = W_L \cdot \ldots \cdot W_1 \bx$, where $\bx \in \R^d$.  
For simplicity, we focus on the binary case with two classes, suggesting that similar qualitative outcomes would apply to the more general multi-class model. Accordingly, we redefine the objective function as follows:
\begin{equation}
\label{eq:object}
\min_{W_1,\ldots,W_L}\sum_{i=1}^n \Vert W_L\cdot\ldots \cdot W_1 \bx_i - y_i \Vert^2  
\end{equation}
Above the matrices $\{W_l\}_{l=1}^L$ represent the $2D$ matrices corresponding to $L$ layers of a deep linear network, and points $\{\bx_i\}_{i=1}^n$ represent the training set with labeling function $y_i=\pm 1$ for the first and second classes, respectively. Note that $\bW = \prod_{l=L}^1 W_l$ is a row vector that defines the resulting separator between the classes. The classifier is defined as:
$f(\bx) = {\text{sign}}~ \left (\prod_{l=L}^1 W_l \bx \right )$ 
for $\bx\in\R^d$.

\paragraph{Preliminaries.}
Let $\bW^{(n)} = \prod_{l=L}^1 W_l^{(n)}$ represent the separator after $n$ GD steps, where $\bW^{(n)} \equiv [w_1^{(n)}, \ldots, w_d^{(n)}] \in \R^d$. For convenience, we rotate the data representation so that its axes align with the eigenvectors of the data's covariance matrix. \citet{hacohen2022principal} showed that the convergence rate of the $j^\mathrm{th}$ element of $\bW$ with respect to $n$ is exponential, governed by the corresponding $j^\mathrm{th}$ eigenvalue:
\begin{equation}
\label{eq:linear-comb}
w_j^{(n)}\approx \lambda_j^{n}w_j^{(0)} + [1- \lambda_j^{n}]w_j^{opt}, \qquad \lambda_j = 1-\gamma s_j L
\end{equation}
Here, $\bW^{(0)}$ denotes the separator at initialization, $\bw^{opt}$ denotes the optimal separator (which can be derived analytically from the objective function), $s_j$ represents the $j^\mathrm{th}$ singular value of the data, and $\gamma$ is the learning rate. Notably, while $\bw^{opt}$ is unique, the specific solution at convergence $\{W_l^{(\infty)}\}_{l=1}^L$ is not.

\subsection{Forget Time in Deep Linear Models}
\label{sec:linear-model-charact}

Let $\Lambda$ denote $\text{diag}(\{\lambda_j\})$ - a diagonal matrix in $\R^{d\times d}$, and I the identity matrix. It follows from (\ref{eq:linear-comb}) that
\begin{equation}
\label{eq:linear-comb-mat}
\bW^{(n)}\approx \bW^{(0)}\Lambda^{n} + \bW^{opt}[I- \Lambda^{n}]
\end{equation}

We say that a point is forgotten if it is classified correctly at initialization, but not so at the end of training. Let $\bx$ denote a forgotten datapoint, and let $N$ denote the number of GD steps at the end of training. Since by definition $f(\bx) = {\text{sign}} (\bW^{(n)} \bx)$, it follows that $\bx$ is forgotten iff 
$\{\bW^{(0)} y\bx > 0\}$ and $\{\bW^{(N)} y\bx < 0\}$. 

Let us define the forget time of point $\bx$ as follows:
\begin{defn}[Forget time]
\label{def:1}
GD iteration $\hat n$ that satisfies 
\begin{equation}
\label{eq:forget-time}
\begin{split}
&\bW^{(\hat n)} y\bx \leq 0 \\
& \bW^{(n)} y\bx > 0 \qquad \forall n<\hat n
\end{split}
\end{equation}
\end{defn}
\begin{claim}
\label{eq:forgt-time}
Each forgotten point has a finite forget time $\hat n$.
\end{claim}
\begin{proof}
Since $\{\bW^{(0)} y\bx > 0\}$ and $\{\bW^{(N)} y\bx < 0\}$, (\ref{eq:forget-time}) follows by induction.
\end{proof}

Note that Def~\ref{def:1} corresponds with the \emph{Forget time} seen in deep networks (cf. Fig.~\ref{subfig:forgettimetimg}). The empirical investigation of this correspondence is discussed in \app\ref{app:linearforget} (see Fig.~\ref{fig:pclinearclassifiersSCLRGD}).


To characterize the time at which a point is forgotten, we inspect the rate with which $F (n)=\bW^{(n)} y\bx $ changes with $n$. We begin by assuming that the learning rate $\gamma$ is infinitesimal, so that terms of magnitude $O(\gamma^2)$ can be neglected. Using (\ref{eq:linear-comb-mat}) and the Taylor expansion of $\lambda_j$ from (\ref{eq:linear-comb})
\begin{equation*}
\begin{split}
F(n) \approx & \left ( \bW^{(0)} - \bW^{opt}  \right ) \Lambda^{n} y \bx + \bW^{opt} y \bx\\
=&  ~\bW^{opt} y \bx + \sum_{j=1}^d ( w_j^{(0)} - w_j^{opt}  ) \lambda_j^{n} y x_j \\
=&  ~\bW^{opt} y \bx + \sum_{j=1}^d ( w_j^{(0)} - w_j^{opt}  ) [\mathsmaller{\mathsmaller{1-n\gamma s_j L + O(\gamma^2) }}] y x_j \\
=&  ~ \bW^{(0)} y \bx - n  \gamma L\sum_{j=1}^d ( w_j^{(0)} - w_j^{opt} ) y s_j x_j + O(\gamma^2) 
\end{split}
\end{equation*}
It follows that
\begin{equation}
\label{eq:forget-time-diff}
\frac{d F(n) }{d n} = -\gamma  y L\sum_{j=1}^d ( w_j^{(0)} - w_j^{opt} ) s_j x_j + O(\gamma^2)
\end{equation}

\noindent
\textbf{Discussion.} 
Recall that $\{s_j\}$ is the set of singular values, ordered such that $s_1 \geq s_2 \geq \dots \geq s_d$, and $x_j$ is the projection of point $\bx$ onto the $j^{\text{th}}$ eigenvector. From (\ref{eq:forget-time-diff}), the rate at which a point is forgotten, if at all, depends on vector $[s_j x_j]_j$, in addition to the random vector $\bW^{(0)} - \bW^{\text{opt}}$ and label $y$. All else being equal, a point will be forgotten faster if the length of its spectral decomposition vector $[x_j]$ is dominated by its first  components, indicating that most of its mass is concentrated in the leading principal components.

\subsection{Spectral Properties of Forgotten Images}

When working with datasets of natural images, where it has been shown that the singular values decrease rapidly at an approximately exponential rate \citep{hyvarinen2009natural}, the role of the singular values becomes even more pronounced. \citet{hacohen2022principal} argued that in the limiting case, the components of the separating hyperplane $\bW^{opt}$ will be learned sequentially, one at a time. In essence, the model first captures the projection of $\bW^{opt}$ onto the data's leading eigenvector, then onto the subsequent eigenvectors in order. For similar considerations, this reasoning also holds in the multi-class scenario. 

This analysis suggests that PCA of the raw data governs the learning  of the linear separator. We therefore hypothesize that forgotten points with substantial projections onto the leading principal components are more likely to be forgotten early, and vice versa. To empirically test this prediction, we must first establish some key definitions.

Let $\hW^{opt}\in\R^{c\times d}$ denote the optimal solution of the multi-class linear model with $c$ classes and the $L_2$ loss. Let $\hW(k)$ denote the projection of $\hW^{opt}$ on the first $k$ principal components of the raw data.
\begin{defn}
\label{def:spectral}
Let $\Sc(k)$ denote the set of points that are correctly classified by $\hW(k')$ for some $k' > k$, but incorrectly classified by $\hW^{opt}$. Similarly, let $\Mc(n)$ denote the set of points correctly classified by the trained deep model after $n' > n$ epochs, but incorrectly classified by the final model.
\end{defn}



\begin{figure}[htbp]
   \centering
    \begin{subfigure}[b]{0.42\linewidth}
        \includegraphics[width=\linewidth]{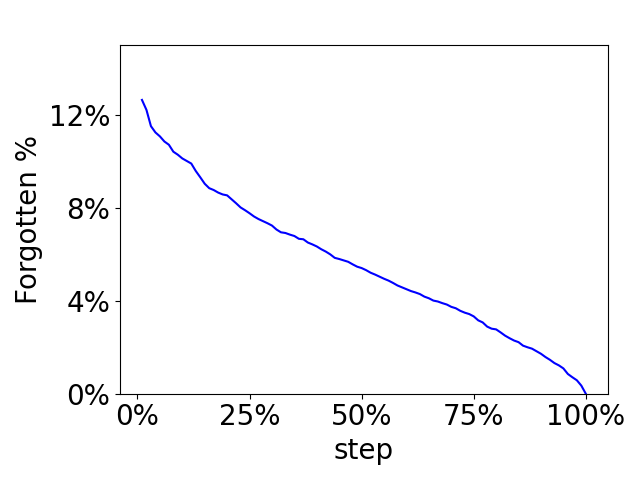}
    \vspace{-.7cm}
        \caption{$\vert \Sc(k)\vert $}
        \label{subfig:FRRGB}
    \end{subfigure}
    \begin{subfigure}[b]{0.42\linewidth}
        \includegraphics[width=\linewidth]{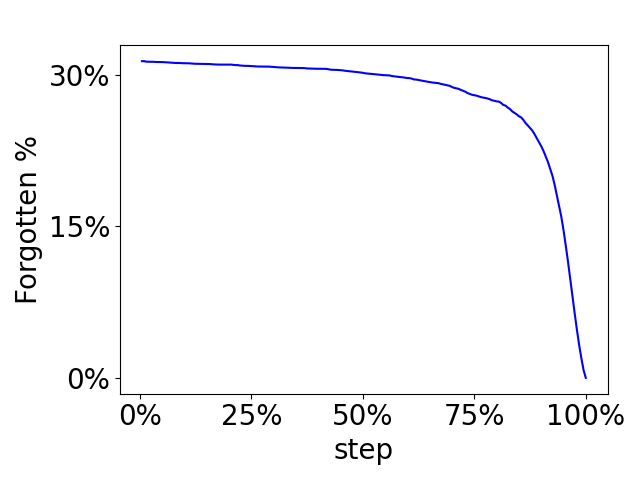}
    \vspace{-.7cm}
        \caption{$\vert \Mc(n)\vert $}
    \end{subfigure}
    \begin{subfigure}[b]{0.42\linewidth}
        \includegraphics[width=\linewidth]{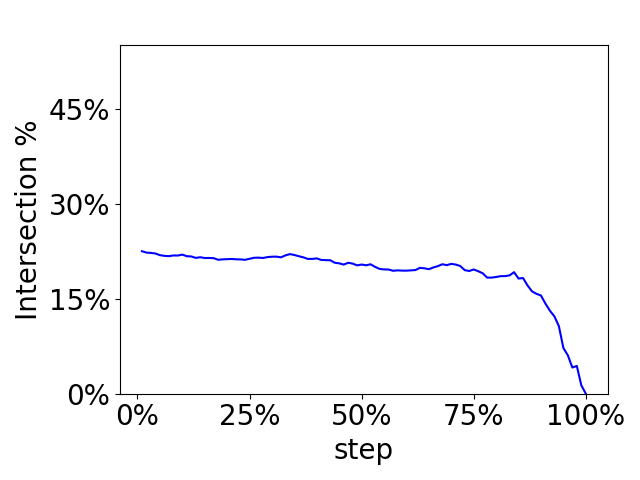}
    \vspace{-.6cm}
        \caption{\normalsize$\frac{\vert \Sc(k)\cap \Mc(n)\vert }{\vert \Sc(k)\vert}$}
        \label{subfig:IRPCRGBITG}
    \end{subfigure}
    \begin{subfigure}[b]{0.42\linewidth}
        \includegraphics[width=\linewidth]{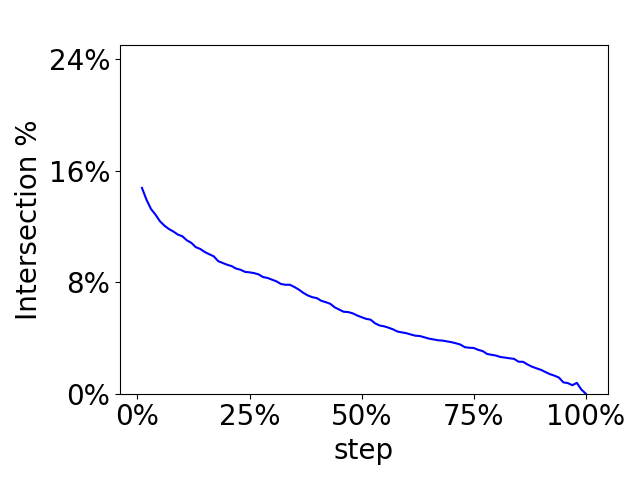}
    \vspace{-.6cm}
        \caption{\normalsize$\frac{\vert \Sc(k)\cap \Mc(n)\vert }{\vert \Mc(n)\vert}$}
        \label{subfig:IROrgRGBITG}
    \end{subfigure}
     \caption{Empirical results, correlating the sets of examples forgotten during the training of a DNN and those forgotten during the training of a deep linear network (see text for details). Note in (d) that early on, roughly $\frac{1}{6}$ of the points to be forgotten by our deep model are also forgotten by the deep linear model.}
     
     \label{fig:pclinearclassifiers}
\vspace{-0.3cm}
\end{figure}

To empirically investigate the prediction above, we correlate the two sets $\Sc(k)$ and $\Mc(n)$ after establishing correspondence $n=\alpha k+\beta$ between the ranges of indices $k$ and $n$. We examined this correlation using the CIFAR100 dataset, a linear model trained using the images' RGB representation, and the corresponding deep model from the experiments reported in Section~\ref{sec:empirical}. Interestingly, the respective sets $\Sc(k)$ and $\Mc(n)$ show significant correlation, as seen in Fig.~\ref{fig:pclinearclassifiers}. Since deep networks also learn a representation, we repeated the experiment with alternative learned feature spaces, obtaining similar results (see Fig.~\ref{fig:lastforgottenLinear} in \app\ref{app:linearforget}). 

\section{Recover Forgotten Knowledge: Algorithm}
\label{sec:method}

In Section~\ref{sec:overfitanddoubledescent} we showed that neural networks often achieve better performance in mid-training on a subset of the test data, even when the test accuracy is monotonically increasing with training epochs. Here we aim to integrate the knowledge obtained in mid- and post-training epochs, during inference time, in order to improve performance. To this end we must determine: \begin{inparaenum}[(i)] \item  which versions of the model to use; \item how to combine them with the post-training model; and \item how to weigh each model in the final ensemble. \end{inparaenum}

\myparagraph{Choosing an early epoch of the network.} Given a set of epochs $\{1,\ldots,E\}$ and corresponding forget rates $\{F_e\}_e$, we first single out the model $n_A$ obtained at epoch $A = argmax_{e \in \{1,...,E\}}F_e$. This epoch is most likely to correctly fix mistakes of the model on "forgotten" test data.

\myparagraph{Combining the predictors.} Next, using validation data we determine the relative weights of the two models - the final model $n_E$, and the intermediate model $n_A$ with maximal forget fraction. Since the accuracy of $n_E$ is typically much higher than $n_A$, and in order not to harm the ensemble's performance, we expect to assign $n_E$ a higher weight. 

\myparagraph{Improving robustness.} To improve our method's robustness to the choice of epoch $A$, we use a window of epochs around $A$, denoted by $\{n_{A-w},...,n_A,..., n_{A+w}\}$. The vectors of probabilities computed by each checkpoint are averaged before forming an ensemble with $n_E$. In our experiments, we use a fixed window $w=1$, achieving close to optimal results as verified in the ablation study (see \app\ref{abl:window}).

\myparagraph{Iterative selection of models.} As we now have a new predictor, we can find another alternative predictor from the training history that maximizes accuracy on the data misclassified by the new predictor, in order to combine their knowledge as described. This can be repeated iteratively, until no further improvement is achieved. 

\myparagraph{Choosing hyper-parameters.} In order to compute $F_e$ and assign optimal model weights and window size, we use a validation set, which is a part of the labeled data not shown to the model during initial training. This is done \textbf{post training} as it has no influence over the training process, and thus \emph{doesn't incur additional training cost}. We follow common practice, and show in \app\ref{abl:val} that after optimizing these hyper-parameters, it is possible to retrain the model on the complete training set while maintaining the same hyper-parameters. The performance of our method thus trained is superior to alternative methods trained on the same data. 

\myparagraph{Pseudo-code for our method.} We name our method \textbf{K}nowledge\textbf{F}usion (KF), and provide its pseudo-code in Alg~\ref{alg:test}. There, we call functions that: (i) calculate the forget value per epoch on some validation data, given the predictions at each epoch (\textbf{calc\_early\_forget}); and (ii) calculate the probability of each class for a given example and a list of predictors (\textbf{get\_class\_probs}). 

\begin{algorithm}[htbp]
\small
   \caption{Knowledge Fusion (KF)}
\begin{algorithmic}
   \STATE {\bfseries Input:} {Checkpoints of trained model \{$n_0$,...,$n_E$\}, w, test-pt $x$}
    \STATE{\bfseries Output:}  prediction for $x$
    \STATE{\{$A_1$,...,$A_k$\},  \{ $\varepsilon_1,...,\varepsilon_k$\} $\gets$ \textbf{calc\_early\_forget}(\{ $n_0$,...,$n_E$\})}
    \STATE{$prob \gets \mathbf{get\_class\_probs}[E]$}\; 
    \FOR{$i\gets1$ {\bfseries to} $k$}
    \STATE{$prob_A \gets mean(\mathbf{get\_class\_probs}[A_i-w:A_i+w])$}\;
    \STATE{$prob \gets \varepsilon_i*prob_A+(1-\varepsilon_i)*prob$}\;
    \ENDFOR
    \STATE{$prediction \gets \mathbf{argmax}(prob)$}\; 

    \STATE{\textbf{Return} $prediction$}
\end{algorithmic}
\label{alg:test}
\end{algorithm}

\myparagraph{Self-distillation post-processing.}
The proposed method enhances the performance of any trained model $m$ with only a minor increase in training costs. However, ensemble classifiers often incur high inference costs. To address this, self-distillation can be employed with a further increase in training costs, to deliver a single model $m'$ that \emph{achieves performance comparable to the ensemble while maintaining inference costs comparable to the original model}.

\section{Empirical Evaluation}
\label{sec:empirical}

\subsection{Main Results}

In this section we evaluate the performance of our method as compared to the original predictor, i.e. the network after training, and other baselines. We use various image classification datasets, neural network architectures, and training schemes. The main results are presented in Tables~\ref{table:regularnetworks}-\ref{table:specialmethods}, followed by a brief review of our extensive ablation study and additional comparisons in Section~\ref{sec:ablation}. 

Specifically, in Table~\ref{table:regularnetworks} we report results while using multiple architectures trained on CIFAR-100, TinyImagenet and Imagenet, with different learning rate schedulers and optimizers. For comparison, we report the results of both the original predictor and some baselines. Additional results for scenarios  connected to overfitting are shown in Table~\ref{table:labelnoise} and \app\ref{app:additionaleval}, where we test our method on these datasets with injected symmetric and asymmetric label noise (see \app\ref{app:implementationdetails}), as well as on a real label noise dataset (Animal10N). Note that, as customary, the label noise exists only in the train data while the test data remains clean for model evaluation. 

\begin{table*}[thb!]
\footnotesize
  \centering
  \begin{tabular}{l| c||c||c|c|c|c}
    \multicolumn{1}{ l |}{Method/\textbf{Dataset}} & \multicolumn{1}{ c ||}{\textbf{CIFAR-100}} & \multicolumn{1}{ c || }{\textbf{TinyImagenet}} & \multicolumn{4}{ c }{\textbf{Imagenet}} 
    \\ 
    \multicolumn{1}{ r |}{architecture} &     Resnet18 & Resnet18 & Resnet50  &  ConvNeXt large & ViT16 base & MaxViT tiny \\
    \toprule
        \emph{single network}   & 
        $78.07 \pm .28$ & $64.95 \pm .24$ & $75.74 \pm .14$ & $82.92 \pm .11$ & $79.16 \pm .1$ & $82.51 \pm .15 $ \\
        \hline
        \emph{horizontal (i)}   & $78.15 \pm .17$ & $64.89 \pm .18$ & $\mathbf{76.46 \pm .14}$ & $\mathbf{83.13 \pm .1}$ & $79.11 \pm .1$ & $82.77 \pm .1$ \\
        \emph{fixed jumps (i)}   & $78.04 \pm .23$ & $ 66.54 \pm .35$ & $75.5 \pm .09$ & $82.37 \pm .1$ & $78.67 \pm .08$ & $\mathbf{83.38 \pm .1}$ \\
        \emph{KF (ours) (i)}   & $\mathbf{78.33 \pm .08}$ & $\mathbf{66.98 \pm .37}$ & $75.88 \pm .14$ & $\mathbf{83.18 \pm .16}$ & $\mathbf{79.93 \pm .11}$ & $\mathbf{83.34 \pm .04}$ \\

        \hline
        \emph{horizontal $(\infty)$}   & $78.23 \pm .17$ & $65.11 \pm .3$ & $\mathbf{76.42 \pm .1}$ & $83.02 \pm .06$ & $79.53 \pm .13$& $82.93 \pm .14$  \\
        \emph{fixed jumps $(\infty)$}   & $\mathbf{79.17 \pm .08}$ & $68.24 \pm .38$ & $75.72 \pm .18$ & $\mathbf{83.86 \pm .06}$ & $79.11 \pm .13$ & $\mathbf{83.78 \pm .15}$ \\

        \emph{KF (ours) $(\infty)$} & $\mathbf{79.13 \pm .14}$ &$\mathbf{68.5 \pm .36}$ & $\mathbf{76.52 \pm .16}$ & $\mathbf{83.96 \pm .09}$ & $\mathbf{80.34 \pm .08}$ & $\mathbf{83.81 \pm .14}$\\
        \bottomrule
        \emph{improvement} & $\mathbf{1.05 \pm .14}$ &$\mathbf{3.54 \pm .14}$ & $\mathbf{.78 \pm .04}$ & $\mathbf{1.03 \pm 13}$ & $\mathbf{1.17 \pm .08}$ & $\mathbf{1.29 \pm .02}$\\      
  \end{tabular}

  \caption{Mean (over random validation/test splits) test accuracy (in percent) and standard error on image classification datasets, comparing our method and  baselines described in the text. The last row shows the improvement of the best performer over the single network. Suffixes: $(i)$ denotes a limited budget scenario, in which we use our method in a non-iterative manner; $(\infty)$ denotes the unlimited budget scenario, where we use our full iterative version. In each case, the baselines employ the same  number of checkpoints as our method. } 
  \label{table:regularnetworks}
\end{table*}

\begin{table*}[thb!]
\footnotesize
  \centering
  \begin{tabular}{l| c|| c|c || c|c || c|c}
    \multicolumn{1}{ c |}{Method/\textbf{Dataset}} & \multicolumn{1}{ c ||}{\textbf{Animal10N}} & \multicolumn{2}{ c ||}{\textbf{CIFAR-100 asym}} & \multicolumn{2}{ c || }{\textbf{CIFAR-100 sym}} & \multicolumn{2}{ c }{\textbf{TinyImagenet}} 
    \\ 
    \multicolumn{1}{ r |}{\% label noise} &    8\% & 20\% & 40\%  &  20\% & 40\% &  20\% & 40\%\\
    \toprule
    \emph{single network}   &  $85.9\pm.3$ & $67.1 \pm .5$ & $49.4 \pm .3$& $65.4 \pm .3$ & $56.9 \pm .1$ &  $56.2 \pm .2$ & $49.8 \pm .3$ \\

        \hline
    \emph{fixed jumps $(\infty)$}   & $87.1\pm.4$ & $73.9 \pm .1$ & $59.9 \pm .6$& $72.8 \pm .1$ & $66.5 \pm .1$ & $60.0 \pm .8$ & $54.16 \pm .3$ \\
    \emph{horizontal $(\infty)$}  & $86.3\pm.3$ & $73.4 \pm .1$ & $58.5 \pm .1$& $71.1 \pm .38$ & $65.2 \pm .1$ & $59.3 \pm .3$ & $51.7  \pm .2$ \\
    \emph{KF (ours) $(\infty)$} & $\mathbf{87.8\pm.4}$ & $\mathbf{74.2 \pm .1}$& $\mathbf{62.1 \pm .5}$& $\mathbf{72.8 \pm .1}$ & $\mathbf{67.0 \pm .1}$ & $\mathbf{62.8 \pm .2}$ &$\mathbf{57.0 \pm .5}$ \\
    \bottomrule
    \emph{improvement} & $ \mathbf{1.9\pm.4}$ & $\mathbf{7.1 \pm .6}$& $\mathbf{12.6 \pm .2}$& $\mathbf{7.4 \pm .4}$ & $\mathbf{10.1 \pm .1}$ & $\mathbf{6.6 \pm .1}$ &$\mathbf{7.2 \pm .1}$ \\

  \end{tabular}
  \caption{Mean test accuracy (in percent) and standard error of Resnet 18, comparing our method and the baselines on datasets with large label noise and significant overfitting. We include a comparison using the Animal10N dataset, which has innate label noise. 
  } 

  \label{table:labelnoise}
\end{table*}

\begin{table*}[thb!]
\footnotesize
  \centering
  \begin{tabular}{l| c || c || c|c || c|c}
    \multicolumn{1}{ c |}{Method/\textbf{Dataset}} & \multicolumn{1}{ c ||}{\textbf{CIFAR-100}}& \multicolumn{1}{ c ||}{\textbf{Animal10N}}& \multicolumn{2}{ c ||}{\textbf{CIFAR-100 asym}} & \multicolumn{2}{ c  }{\textbf{CIFAR-100 sym}} 
    \\ 
    \multicolumn{1}{ r |}{\% label noise} &     0\% & 8\% & 20\% & 40\%  &  20\% & 40\% \\
    \toprule
    \emph{FGE $(\infty)$}   & $78.9 \pm .4$& $86.5\pm0.6$ &$67.1 \pm .2$& $48.1 \pm .3$& $66.5 \pm .1$& $52.1 \pm .1$\\
    \emph{SWA $(\infty)$}   & $78.8 \pm .1$& $\mathbf{88.1\pm.2}$& $66.6 \pm .1$& $46.9 \pm .2$& $65.6 \pm .4$& $50.0 \pm .1$\\
    \emph{snapshot $(\infty)$}   & $78.4 \pm .1$ & $86.8\pm.3$ &  $72.1 \pm .4$ & $52.8 \pm .6$& $70.8 \pm .5$ & $63.8 \pm .2$  \\
    \emph{KF (ours) $(\infty)$} & $\mathbf{79.3 \pm .2}$ & $\mathbf{87.8\pm.4}$& $\mathbf{74.2 \pm .1}$& $\mathbf{62.1 \pm .5}$& $\mathbf{72.8 \pm .1}$ & $\mathbf{67.0 \pm .1}$\\
  \end{tabular}
  \caption{Mean test accuracy of Resnet18, using for baseline methods that alter the training procedure.} 
  \label{table:specialmethods}
\end{table*}

In Table~\ref{table:specialmethods} and \app\ref{app:additionaleval} we compare our method to additional methods that adjust the training protocol itself, using both clean and noisy datasets. We employ these methods using the same network architecture as our own, after a suitable hyper-parameter tuning.

In each experiment we use half of the \emph{test data} for validation, to compute our method's hyper-parameters (the list of alternative epochs and $\{\varepsilon_i\}$), and then test the result on the remaining test data. The accuracy reported here is only on the remaining test data, averaged over three random splits of validation and test data, using different random seeds. In \app\ref{abl:val} we report results on the original train/test split, where a subset of the training data is set aside for hyper-parameter tuning.  As customary, these same parameters are later used with models trained on the full training set, demonstratively without deteriorating the results.

\myparagraph{Baselines} Our method incurs the training cost of a single model, and thus, following the methodology of \citep{huang2017snapshot}, we compare ourselves to methods that require the same amount of training time. The first group of baselines includes methods that do not alter the training process:
\begin{itemize}[leftmargin=0.25cm,topsep=0pt]
\setlength\itemsep{-0.15em}
    \item \textbf{Single network}: the original network, after training. 
    \item \textbf{Horizontal ensemble} \citep{xie2013horizontal}: this method uses a set of epochs at the end of the training, and delivers their average probability outputs (with the same number of checkpoints as we do). 
    \item \textbf{Fixed jumps}: this baseline was used in \citep{huang2017snapshot}, where several checkpoints of the network, equally spaced through time, are taken as an ensemble.
\end{itemize} 

The second group includes methods that \emph{alter} the training protocol. While this is not a directly comparable set of methods, as they focus on a complementary way to improve performance, we report their results in order to further validate the usefulness of our method. This group includes \textit{Snapshot ensemble} \citep{huang2017snapshot}, \textit{Stochastic Weight Averaging} (SWA) \citep{izmailov2018averaging} and \textit{Fast Geometric Ensembling} (FGE) \citep{garipov2018loss}, see details in \app\ref{app:base}. Comparisons to additional baselines that are relevant to resisting overfitting, including early stopping and test time augmentation, are discussed in \app\ref{abl:ema}. Full implementation details are provided in \app\ref{app:implementationdetails}.

\subsection{Ablation Study}
\label{sec:ablation}

We conducted an extensive ablation study in order to investigate the limitations, and some practical aspects, of our method. Due to space limitation, we only provide here a brief overview of the results, and postpone the full description to \app\ref{app:additionalablation}. The results can be summarized as follows:

\begin{inparaenum}[(i)] \item \S\ref{abl:val} shows that a separate validation set is not really necessary for the method to work well. \item \S\ref{subsec:checkpointsnum} investigates how many checkpoints are needed for the method to be effective, showing that only $5-10\%$ of the past checkpoints are sufficient. \item \S\ref{abl:subopttrain} investigates the added value of our method when using only a partial hyper-parameter search, which leads to sub-optimal training. Interestingly, our method is shown to be even more beneficial in the sub-optimal regime, with a smaller gap between the optimal and sub-optimal networks. \item \S\ref{abl:transferlearning} shows that our method is effective in a transfer learning scenario, when using a pre-trained network. \item \S\ref{abl:ema} shows that our method outperforms Exponential-Moving-Average (EMA), early stopping and test time augmentation. \item \S\ref{subsec:modelsize} shows that our method's benefit increases as the number of parameters grows. \item \S\ref{subsec:regens} shows that much of the improvement of a regular ensemble of independent networks can often be obtained by using our method at a much lower cost. \item \S\ref{abl:fairness} shows that our method does not have negative effects on the model's fairness. \item \S\ref{abl:window} shows that using a window of size w=1 is both necessary and near optimal.\end{inparaenum}

\subsection{Method: Discussion and Limitations}  

Our method significantly improves performance in modern neural networks. It complements other overfitting reduction methods like EMA and proves effective where these methods fail, as in fine-tuning of pre-trained models. The method offers substantial performance gains with minimal overhead in inference and memory costs since overlapping model parts are computed once. In challenging scenarios, such as small networks handling complex data or datasets with label noise, it further enhances performance, reducing errors by around 15\% in cases of 10\% asymmetric noise. Our approach outperforms or matches baselines, especially in settings like ViT16 over ImageNet and Resnet18 over TinyImageNet, regardless of training choices. Unlike some horizontal methods and fixed-jump schedules that show limited improvement, our method remains effective without extensive hyper-parameter tuning. Despite needing validation data and multiple checkpoints, these limitations can be mitigated, and the method can revert to a single network if no validation set improvement is observed. 

\section{Summary and Conclusions}

We revisited the problem of \emph{overfitting} in deep learning, proposing to track the forgetting of validation data in order to detect local overfitting. We connected our new perspective with the \emph{epoch wise double descent} phenomenon, empirically extending its scope while demonstrating that a similar effect occurs in benchmark datasets with clean labels. Inspired by these new empirical observations, we constructed a simple yet general method to improve classification at inference time. We then empirically demonstrated its effectiveness on many datasets and modern network architectures. The method improves modern networks by around 1\% accuracy over Imagenet, and is especially useful in some transfer learning settings where its benefit is large and its overhead is very small. Most importantly, the success of the method to improve upon the original model shows that indeed models forget useful knowledge at the late stages of learning, and serves as a proof of concept that recovering this knowledge can be useful to improve performance. 


\section{Acknowledgments}
This work was supported by grants from the Israeli Council of Higher Education and the Gatsby Charitable Foundation.



\bibliography{bib}

\begin{thebibliography}{40}
\providecommand{\natexlab}[1]{#1}

\bibitem[{Allen-Zhu and Li(2023)}]{allenzhu2023}
Allen-Zhu, Z.; and Li, Y. 2023.
\newblock Towards Understanding Ensemble, Knowledge Distillation and Self-Distillation in Deep Learning.
\newblock arXiv:2012.09816.

\bibitem[{Annavarapu(2021)}]{annavarapu2021deep}
Annavarapu, C. S.~R. 2021.
\newblock Deep learning-based improved snapshot ensemble technique for COVID-19 chest X-ray classification.
\newblock \emph{Applied Intelligence}, 51: 3104--3120.

\bibitem[{Arora et~al.(2019)Arora, Cohen, Golowich, and Hu}]{DBLP:conf/iclr/AroraCGH19}
Arora, S.; Cohen, N.; Golowich, N.; and Hu, W. 2019.
\newblock A Convergence Analysis of Gradient Descent for Deep Linear Neural Networks.
\newblock In \emph{7th International Conference on Learning Representations, {ICLR} 2019, New Orleans, LA, USA, May 6-9, 2019}. OpenReview.net.

\bibitem[{Arora, Cohen, and Hazan(2018)}]{arora2018optimization}
Arora, S.; Cohen, N.; and Hazan, E. 2018.
\newblock On the Optimization of Deep Networks: Implicit Acceleration by Overparameterization.
\newblock In \emph{International Conference on Machine Learning}, 244--253.

\bibitem[{Arpit et~al.(2017)Arpit, Jastrzebski, Ballas, Krueger, Bengio, Kanwal, Maharaj, Fischer, Courville, Bengio et~al.}]{arpit2017closer}
Arpit, D.; Jastrzebski, S.; Ballas, N.; Krueger, D.; Bengio, E.; Kanwal, M.~S.; Maharaj, T.; Fischer, A.; Courville, A.; Bengio, Y.; et~al. 2017.
\newblock A closer look at memorization in deep networks.
\newblock In \emph{International conference on machine learning}, 233--242. PMLR.

\bibitem[{Belkin et~al.(2019)Belkin, Hsu, Ma, and Mandal}]{belkin2019reconciling}
Belkin, M.; Hsu, D.; Ma, S.; and Mandal, S. 2019.
\newblock Reconciling modern machine-learning practice and the classical bias--variance trade-off.
\newblock \emph{Proceedings of the National Academy of Sciences}, 116(32): 15849--15854.

\bibitem[{Chen et~al.(2020)Chen, Kornblith, Norouzi, and Hinton}]{chen2020simple}
Chen, T.; Kornblith, S.; Norouzi, M.; and Hinton, G. 2020.
\newblock A simple framework for contrastive learning of visual representations.
\newblock In \emph{International conference on machine learning}, 1597--1607. PMLR.

\bibitem[{Deng et~al.(2009)Deng, Dong, Socher, Li, Li, and Fei-Fei}]{deng2009imagenet}
Deng, J.; Dong, W.; Socher, R.; Li, L.-J.; Li, K.; and Fei-Fei, L. 2009.
\newblock Imagenet: A large-scale hierarchical image database.
\newblock In \emph{2009 IEEE conference on computer vision and pattern recognition}, 248--255. Ieee.

\bibitem[{Dosovitskiy et~al.(2020)Dosovitskiy, Beyer, Kolesnikov, Weissenborn, Zhai, Unterthiner, Dehghani, Minderer, Heigold, Gelly et~al.}]{dosoViTskiy2020image}
Dosovitskiy, A.; Beyer, L.; Kolesnikov, A.; Weissenborn, D.; Zhai, X.; Unterthiner, T.; Dehghani, M.; Minderer, M.; Heigold, G.; Gelly, S.; et~al. 2020.
\newblock An image is worth 16x16 words: Transformers for image recognition at scale.
\newblock \emph{arXiv preprint arXiv:2010.11929}.

\bibitem[{Fukumizu(1998)}]{fukumizu1998effect}
Fukumizu, K. 1998.
\newblock Effect of batch learning in multilayer neural networks.
\newblock \emph{Gen}, 1(04): 1E--03.

\bibitem[{Ganaie et~al.(2022)Ganaie, Hu, Malik, Tanveer, and Suganthan}]{ganaie2022ensemble}
Ganaie, M.~A.; Hu, M.; Malik, A.; Tanveer, M.; and Suganthan, P. 2022.
\newblock Ensemble deep learning: A review.
\newblock \emph{Engineering Applications of Artificial Intelligence}, 115: 105151.

\bibitem[{Garipov et~al.(2018)Garipov, Izmailov, Podoprikhin, Vetrov, and Wilson}]{garipov2018loss}
Garipov, T.; Izmailov, P.; Podoprikhin, D.; Vetrov, D.~P.; and Wilson, A.~G. 2018.
\newblock Loss surfaces, mode connectivity, and fast ensembling of dnns.
\newblock \emph{Advances in neural information processing systems}, 31.

\bibitem[{Guo, Jin, and Liu(2023)}]{guo2023stochastic}
Guo, H.; Jin, J.; and Liu, B. 2023.
\newblock Stochastic weight averaging revisited.
\newblock \emph{Applied Sciences}, 13(5): 2935.

\bibitem[{Hacohen and Weinshall(2022)}]{hacohen2022principal}
Hacohen, G.; and Weinshall, D. 2022.
\newblock Principal components bias in over-parameterized linear models, and its manifestation in deep neural networks.
\newblock \emph{Journal of Machine Learning Research}, 23(155): 1--46.

\bibitem[{He et~al.(2016)He, Zhang, Ren, and Sun}]{he2016deep}
He, K.; Zhang, X.; Ren, S.; and Sun, J. 2016.
\newblock Deep residual learning for image recognition.
\newblock In \emph{Proceedings of the IEEE conference on computer vision and pattern recognition}, 770--778.

\bibitem[{Heckel and Yilmaz(2020)}]{heckel2020early}
Heckel, R.; and Yilmaz, F.~F. 2020.
\newblock Early stopping in deep networks: Double descent and how to eliminate it.
\newblock \emph{arXiv preprint arXiv:2007.10099}.

\bibitem[{Hu, Xiao, and Pennington(2020)}]{DBLP:conf/iclr/HuXP20}
Hu, W.; Xiao, L.; and Pennington, J. 2020.
\newblock Provable Benefit of Orthogonal Initialization in Optimizing Deep Linear Networks.
\newblock In \emph{8th International Conference on Learning Representations, {ICLR} 2020, Addis Ababa, Ethiopia, April 26-30, 2020}. OpenReview.net.

\bibitem[{Huang et~al.(2017)Huang, Li, Pleiss, Liu, Hopcroft, and Weinberger}]{huang2017snapshot}
Huang, G.; Li, Y.; Pleiss, G.; Liu, Z.; Hopcroft, J.~E.; and Weinberger, K.~Q. 2017.
\newblock Snapshot ensembles: Train 1, get m for free.
\newblock \emph{arXiv preprint arXiv:1704.00109}.

\bibitem[{Hyv{\"a}rinen, Hurri, and Hoyer(2009)}]{hyvarinen2009natural}
Hyv{\"a}rinen, A.; Hurri, J.; and Hoyer, P.~O. 2009.
\newblock \emph{Natural image statistics: A probabilistic approach to early computational vision.}, volume~39.
\newblock Springer Science \& Business Media.

\bibitem[{Izmailov et~al.(2018)Izmailov, Podoprikhin, Garipov, Vetrov, and Wilson}]{izmailov2018averaging}
Izmailov, P.; Podoprikhin, D.; Garipov, T.; Vetrov, D.; and Wilson, A.~G. 2018.
\newblock Averaging weights leads to wider optima and better generalization.
\newblock \emph{arXiv preprint arXiv:1803.05407}.

\bibitem[{Jeong and Chung(2024)}]{jeong2024understanding}
Jeong, H.; and Chung, H.~W. 2024.
\newblock Understanding Self-Distillation and Partial Label Learning in Multi-Class Classification with Label Noise.
\newblock \emph{arXiv preprint arXiv:2402.10482}.

\bibitem[{Krizhevsky, Hinton et~al.(2009)}]{krizhevsky2009learning}
Krizhevsky, A.; Hinton, G.; et~al. 2009.
\newblock Learning multiple layers of features from tiny images.

\bibitem[{Le and Yang(2015)}]{le2015tiny}
Le, Y.; and Yang, X. 2015.
\newblock Tiny imagenet visual recognition challenge.
\newblock \emph{CS 231N}, 7(7): 3.

\bibitem[{Liu et~al.(2022)Liu, Mao, Wu, Feichtenhofer, Darrell, and Xie}]{liu2022convnet}
Liu, Z.; Mao, H.; Wu, C.-Y.; Feichtenhofer, C.; Darrell, T.; and Xie, S. 2022.
\newblock A ConvNet for the 2020s.
\newblock \emph{Proceedings of the IEEE/CVF Conference on Computer Vision and Pattern Recognition (CVPR)}.

\bibitem[{McCloskey and Cohen(1989)}]{mccloskey1989catastrophic}
McCloskey, M.; and Cohen, N.~J. 1989.
\newblock Catastrophic interference in connectionist networks: The sequential learning problem.
\newblock In \emph{Psychology of learning and motivation}, volume~24, 109--165. Elsevier.

\bibitem[{Nakkiran et~al.(2021)Nakkiran, Kaplun, Bansal, Yang, Barak, and Sutskever}]{nakkiran2021deep}
Nakkiran, P.; Kaplun, G.; Bansal, Y.; Yang, T.; Barak, B.; and Sutskever, I. 2021.
\newblock Deep double descent: Where bigger models and more data hurt.
\newblock \emph{Journal of Statistical Mechanics: Theory and Experiment}, 2021(12): 124003.

\bibitem[{Noppitak and Surinta(2022)}]{noppitak2022dropcyclic}
Noppitak, S.; and Surinta, O. 2022.
\newblock dropCyclic: snapshot ensemble convolutional neural network based on a new learning rate schedule for land use classification.
\newblock \emph{IEEE Access}, 10: 60725--60737.

\bibitem[{Patrini et~al.(2017)Patrini, Rozza, Krishna~Menon, Nock, and Qu}]{patrini2017making}
Patrini, G.; Rozza, A.; Krishna~Menon, A.; Nock, R.; and Qu, L. 2017.
\newblock Making deep neural networks robust to label noise: A loss correction approach.
\newblock In \emph{Proceedings of the IEEE conference on computer vision and pattern recognition}, 1944--1952.

\bibitem[{Polikar(2012)}]{polikar2012ensemble}
Polikar, R. 2012.
\newblock Ensemble learning.
\newblock \emph{Ensemble machine learning: Methods and applicannavarapu2021deepations}, 1--34.

\bibitem[{Polyak and Juditsky(1992)}]{polyak1992acceleration}
Polyak, B.~T.; and Juditsky, A.~B. 1992.
\newblock Acceleration of stochastic approximation by averaging.
\newblock \emph{SIAM journal on control and optimization}, 30(4): 838--855.

\bibitem[{Saxe, McClelland, and Ganguli(2014)}]{DBLP:journals/corr/SaxeMG13}
Saxe, A.~M.; McClelland, J.~L.; and Ganguli, S. 2014.
\newblock Exact solutions to the nonlinear dynamics of learning in deep linear neural networks.
\newblock In \emph{2nd International Conference on Learning Representations, {ICLR} 2014, Banff, AB, Canada, April 14-16, 2014, Conference Track Proceedings}.

\bibitem[{Srivastava et~al.(2014)Srivastava, Hinton, Krizhevsky, Sutskever, and Salakhutdinov}]{srivastava2014dropout}
Srivastava, N.; Hinton, G.; Krizhevsky, A.; Sutskever, I.; and Salakhutdinov, R. 2014.
\newblock Dropout: a simple way to prevent neural networks from overfitting.
\newblock \emph{The journal of machine learning research}, 15(1): 1929--1958.

\bibitem[{Stephenson and Lee(2021)}]{stephenson2021and}
Stephenson, C.; and Lee, T. 2021.
\newblock When and how epochwise double descent happens.
\newblock \emph{arXiv preprint arXiv:2108.12006}.

\bibitem[{Stern, Shwartz, and Weinshall(2024)}]{stern2024united}
Stern, U.; Shwartz, D.; and Weinshall, D. 2024.
\newblock United We Stand: Using Epoch-wise Agreement of Ensembles to Combat Overfit.
\newblock In \emph{Proceedings of the AAAI Conference on Artificial Intelligence}, volume 38(13), 15075--15082.

\bibitem[{TorchVision(2016)}]{torchvision2016}
TorchVision. 2016.
\newblock TorchVision: PyTorch's Computer Vision library.
\newblock \url{https://github.com/pytorch/vision}.

\bibitem[{Tu et~al.(2022)Tu, Talebi, Zhang, Yang, Milanfar, Bovik, and Li}]{tu2022MaxViT}
Tu, Z.; Talebi, H.; Zhang, H.; Yang, F.; Milanfar, P.; Bovik, A.; and Li, Y. 2022.
\newblock Maxvit: Multi-axis vision transformer.
\newblock In \emph{Computer Vision--ECCV 2022: 17th European Conference, Tel Aviv, Israel, October 23--27, 2022, Proceedings, Part XXIV}, 459--479. Springer.

\bibitem[{Wang et~al.(2020)Wang, Qinami, Karakozis, Genova, Nair, Hata, and Russakovsky}]{wang2020towards}
Wang, Z.; Qinami, K.; Karakozis, I.~C.; Genova, K.; Nair, P.; Hata, K.; and Russakovsky, O. 2020.
\newblock Towards fairness in visual recognition: Effective strategies for bias mitigation.
\newblock In \emph{Proceedings of the IEEE/CVF conference on computer vision and pattern recognition}, 8919--8928.

\bibitem[{Xie, Xu, and Chuang(2013)}]{xie2013horizontal}
Xie, J.; Xu, B.; and Chuang, Z. 2013.
\newblock Horizontal and vertical ensemble with deep representation for classification.
\newblock \emph{arXiv preprint arXiv:1306.2759}.

\bibitem[{Yang, Lv, and Chen(2023)}]{yang2023survey}
Yang, Y.; Lv, H.; and Chen, N. 2023.
\newblock A survey on ensemble learning under the era of deep learning.
\newblock \emph{Artificial Intelligence Review}, 56(6): 5545--5589.

\bibitem[{Zhao et~al.(2017)Zhao, Wang, Yatskar, Ordonez, and Chang}]{zhao2017men}
Zhao, J.; Wang, T.; Yatskar, M.; Ordonez, V.; and Chang, K.-W. 2017.
\newblock Men also like shopping: Reducing gender bias amplification using corpus-level constraints.
\newblock \emph{arXiv preprint arXiv:1707.09457}.

\end{thebibliography}

\newpage

\appendix

\section*{Appendix}


\section{Additional Demonstrations of Forgetting}
\label{app:moreforget}

\begin{figure*}[htbp]
    \centering
    \begin{subfigure}[b]{0.26\linewidth}
        \includegraphics[width=\linewidth]{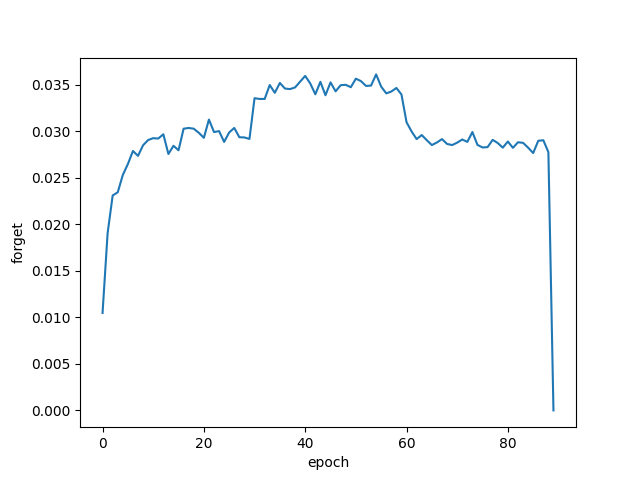}
    \vspace{-.5cm}
        \caption{Imagenet, Resnet50, steplr}
        \label{subfig:r18forgetImg}
    \end{subfigure}
    \begin{subfigure}[b]{0.24\linewidth}
        \includegraphics[width=\linewidth]{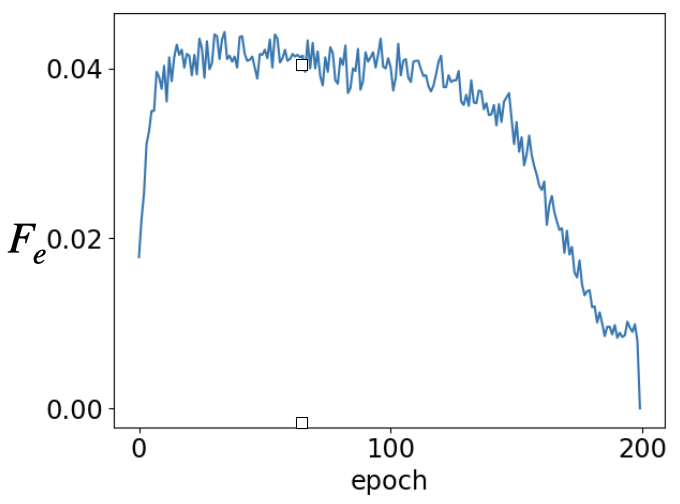}
    \vspace{-.5cm}        \caption{Cifar100, Densenet121}
        \label{subfig:ForgetDenseC100}
    \end{subfigure}
    \begin{subfigure}[b]{0.24\linewidth}
        \includegraphics[width=\linewidth]{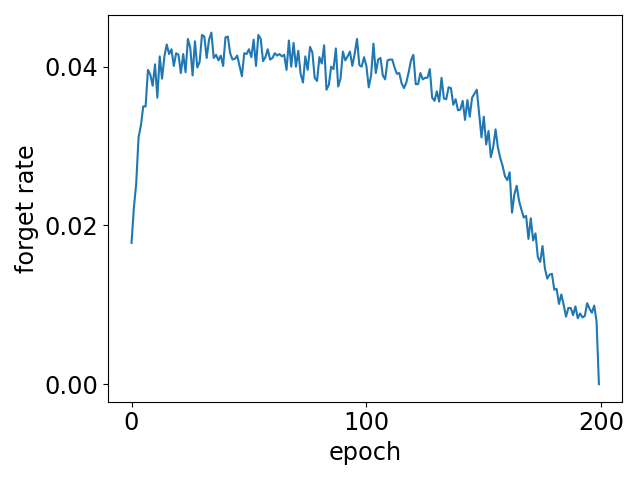}
    \vspace{-.5cm}
        \caption{TinyImagenet, Densenet121}
        \label{subfig:Denseforgettiny}
    \end{subfigure}
    \centering
    \begin{subfigure}[b]{0.24\linewidth}
        \includegraphics[width=\linewidth]{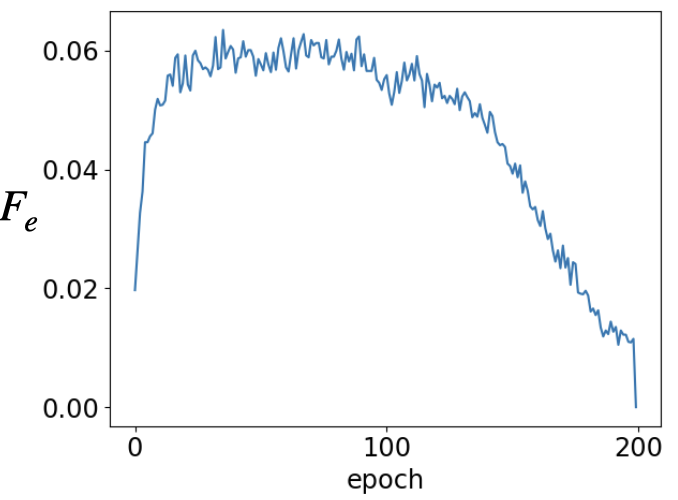}
    \vspace{-.5cm}
        \caption{TinyImagenet, Resnet 18}
    \end{subfigure}
     \caption[forget]{ The forget fraction $F_e$, as defined in (\ref{eq:forget}), of common neural networks trained on various image classification datasets and different architectures. } 
     \label{fig:moreforgetrate}
\end{figure*}

\begin{figure}[tbp]
    \begin{subfigure}[b]{0.325\linewidth}
        \includegraphics[width=\linewidth]{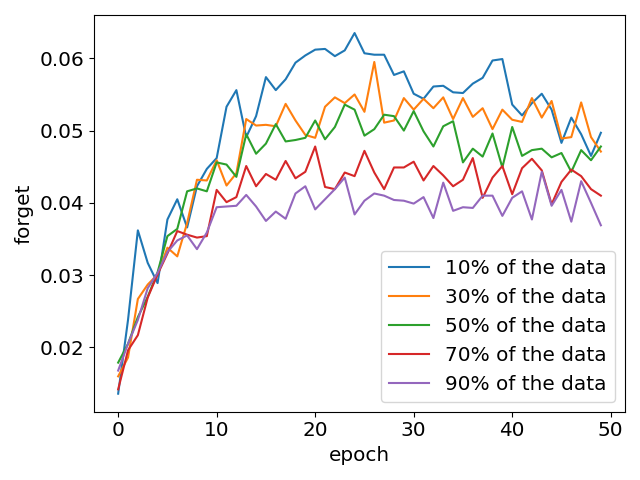}
    \vspace{-.5cm}
        \caption{Cifar100, resnet 18}
    \end{subfigure}
    \begin{subfigure}[b]{0.325\linewidth}
        \includegraphics[width=\linewidth]{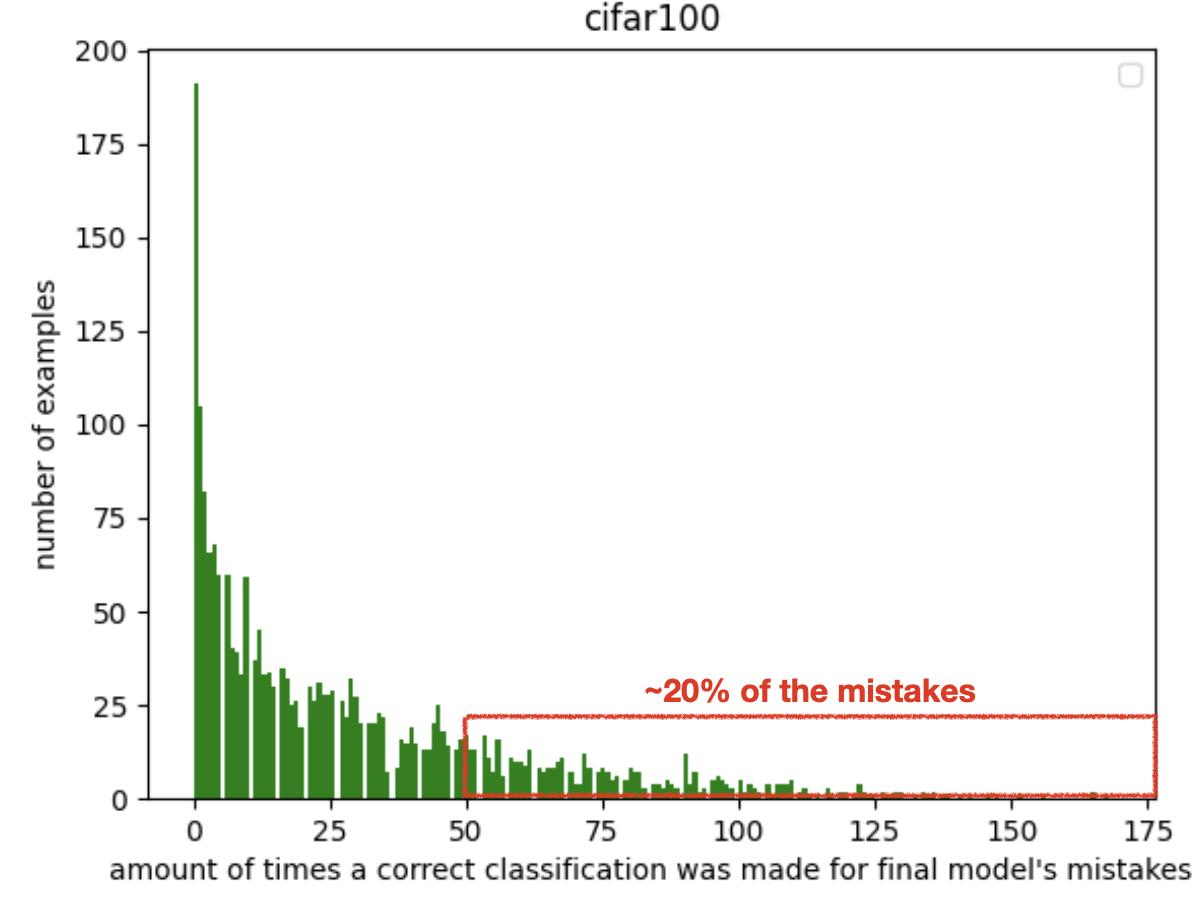}
    \vspace{-.5cm}
        \caption{CIFAR-100, resnet 18}
    \end{subfigure}
    \begin{subfigure}[b]{0.325\linewidth}
        \includegraphics[width=\linewidth]{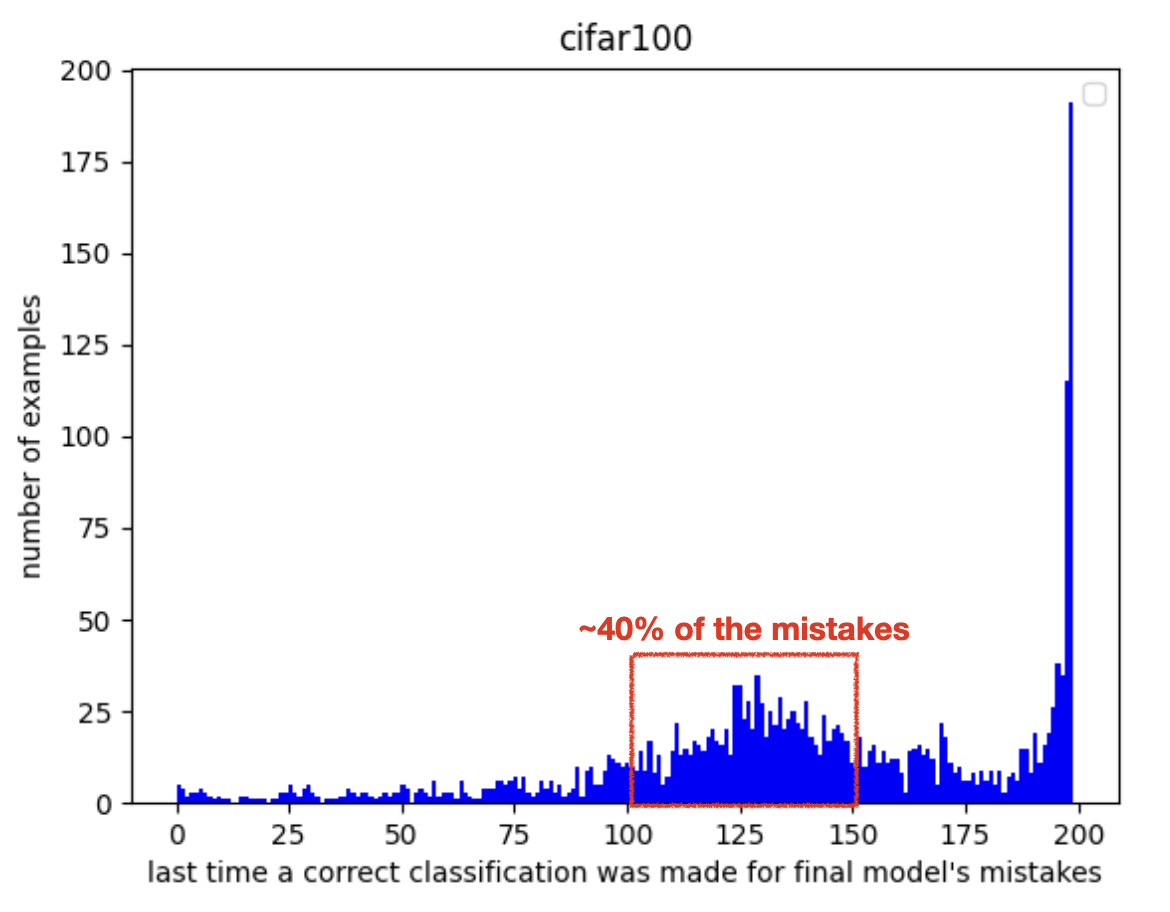}
    \vspace{-.5cm}
        \caption{CIFAR-100, resnet 18}
    \end{subfigure}
     \caption[forget]
     {(a) The $F_e$ score of Resnet18 trained on $10/30/50/70/90\%$ of the train data in CIFAR-100 (purple/red/green/yellow/blue line, respectively) in the first 50 epochs of training (after which the score decreases); $F_e$ is significantly larger for the smallest set of only $10\%$. (b-c) Within the set of wrongly classified test points after training, we show (b) the fraction that was correctly predicted (y-axis) for x epochs (x-axis), and (c) the last epoch in which an example was classified correctly.} 
    \label{fig:forgetrate}
\end{figure}

We first show more examples of various neural networks trained on different datasets which show significant forgetting during training (Fig~\ref{fig:moreforgetrate}), in order to further demonstrate the generality of this phenomenon. 
Additional results on local overfitting with different datasets are shown in Fig.~\ref{fig:forgetrate}.

\section{Forgetting in a Deep Linear Model}
\label{app:linearforget}

\begin{figure*}[htbp]
    \centering
    \begin{subfigure}[b]{0.3\linewidth}
        \includegraphics[width=\linewidth]{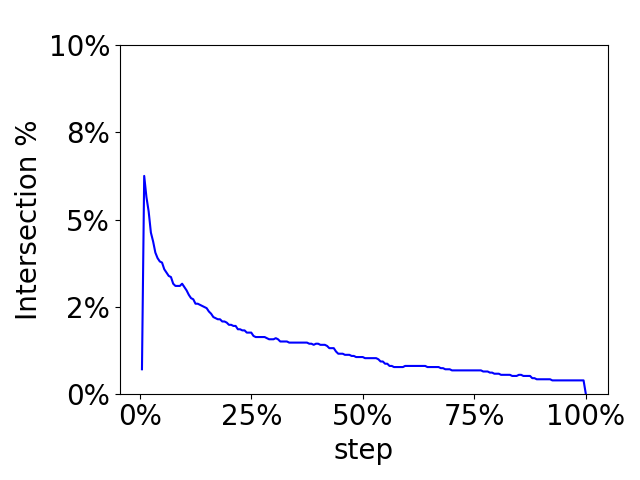}
        \includegraphics[width=\linewidth]{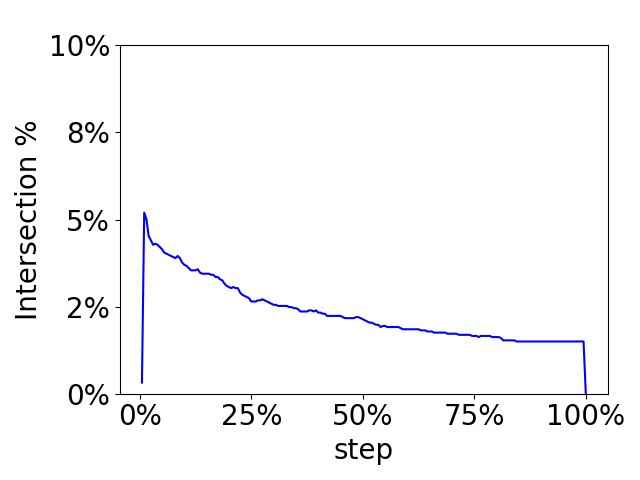}
        \includegraphics[width=\linewidth]{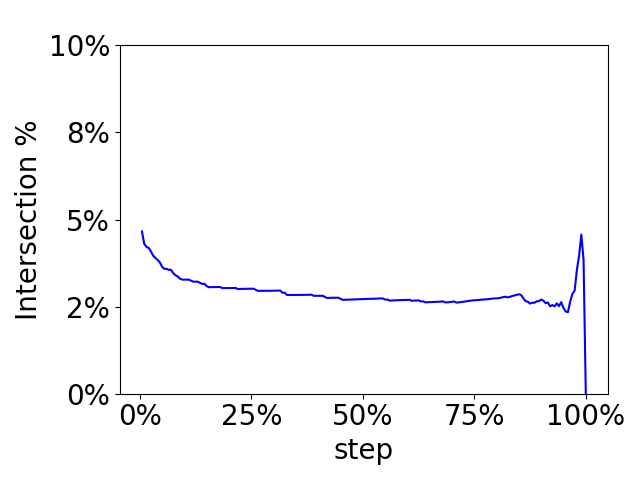}
    \vspace{-.6cm}
        \caption{\normalsize$\frac{\vert F_e^L\cap F_e^D\vert }{\vert F_e^D\vert}$}
        \label{subfig:IROrgTLGD}
    \end{subfigure}
    \begin{subfigure}[b]{0.3\linewidth}
        \includegraphics[width=\linewidth]{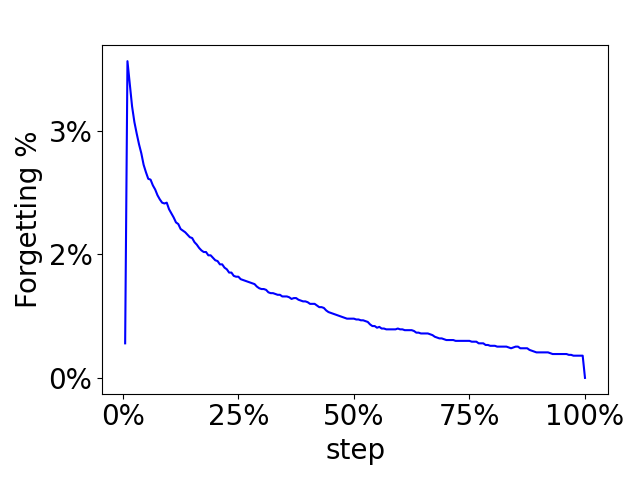}
        \includegraphics[width=\linewidth]{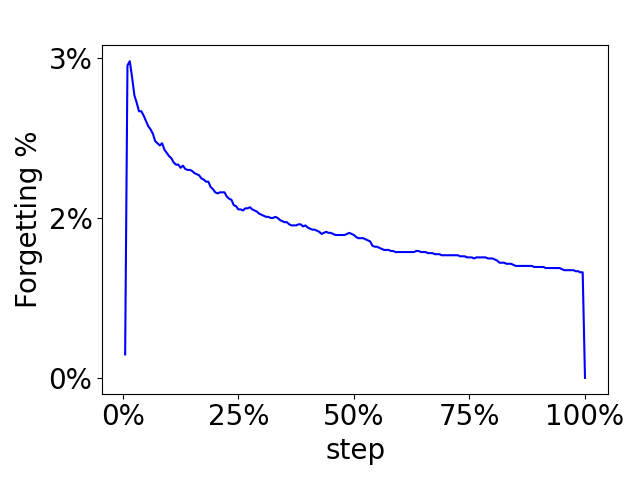}
        \includegraphics[width=\linewidth]{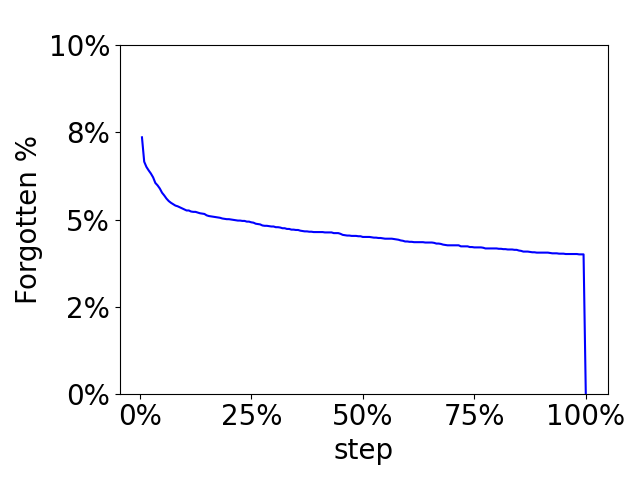}
    \vspace{-.6cm}
        \caption{$\overset{\textcolor{white}{F_e^L}}{F_e^L}$}
        \label{subfig:FRTLGD}
    \end{subfigure}
    \begin{subfigure}[b]{0.3\linewidth}
        \includegraphics[width=\linewidth]{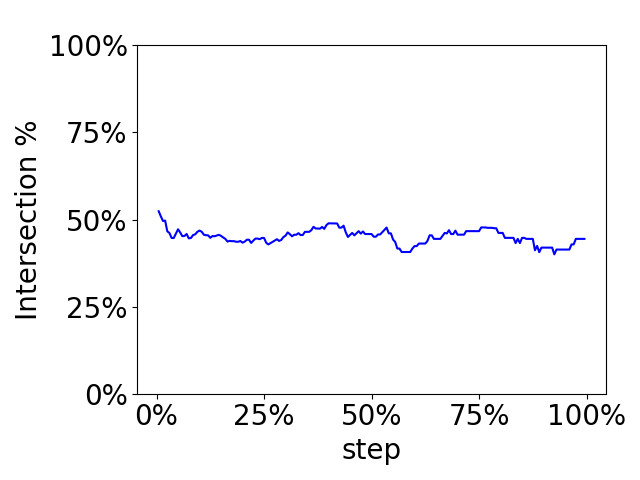}
        \includegraphics[width=\linewidth]{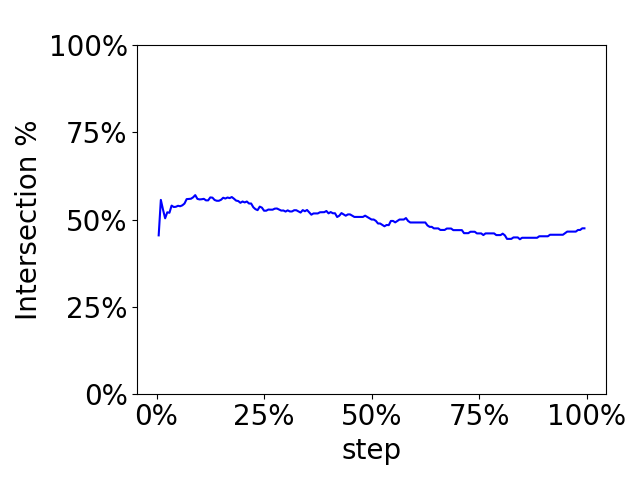}
        \includegraphics[width=\linewidth]{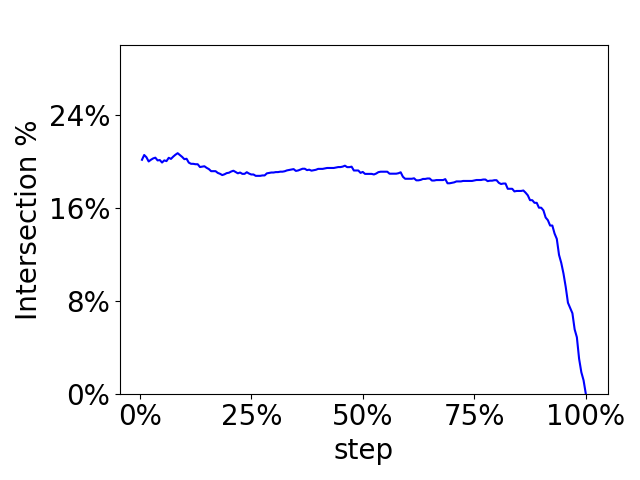}
    \vspace{-.6cm}
        \caption{\normalsize$\frac{\vert F_e^L\cap F_e^D\vert }{\vert F_e^L\vert}$}
        \label{subfig:IRPCTLGD}
    \end{subfigure}

     \caption[heatmap]{Empirical results using dataset CIFAR-100, matching the examples forgotten during the training of a DNN and those forgotten during the training of a deep linear network (see details in text). $F_e^L$ and $F_e^D$ denote the $F_e$ score (\ref{eq:forget}), as a function of epoch $e$, of a deep linear network and a DNN respectively. Three features spaces are used, computed with: top - the self-supervised method described in \citep{chen2020simple}; middle - representation based on transfer learning from ResNet50 trained on ImageNet; bottom - original RGB pictures vectorized;}
     
     \label{fig:pclinearclassifiersSCLRGD}
     \vspace{-.5cm}
\end{figure*}

To investigate the relevance of the theoretical discussion in Section~\ref{sec:linear-model-charact}, we empirically analyze the correlation between test examples forgotten during the training of a deep linear model and those forgotten during the training of a state-of-the-art deep model. We employed ResNet-18 as our deep neural network, alongside a single-layer linear model with a width of $C\times D$, where $C$ represents the number of classes and $D$ denotes the sample dimension. Results are shown in Fig.~\ref{fig:pclinearclassifiersSCLRGD}.

\section{Computing Hyper-Parameters}
\label{app:valalg}

\begin{algorithm}[htb]
   \caption{KF - hyper-parameter calculation}
\begin{algorithmic}
   \STATE {\bfseries Input:}{all past checkpoints during training of the neural network \{$n_0$,...,$n_E$\}, w and validation data $V$, w and validation data $V$
    \STATE{\bfseries Output:}  list of alternative epochs and their weights}
   \STATE $class\_probs \gets$ 
\textbf{get\_class\_probs}(\{$n_0$,...,$n_E$\}, $V$)\;
    \STATE {$prob \gets class\_probs[E]$}\;
    \STATE {explore = \{$n_0$,...,$n_E$\}}\;
    \STATE {Alternative\_epochs = \{\}}\;
    \STATE {epsilons = \{\}}\;
   \WHILE{explore is not empty}
   \STATE {$F$ = \textbf{calc\_forget\_per\_epoch}($prob$, $class\_probs$)}\;
    \STATE {$alt\_epoch$ = \textbf{argmax}($F[explore]$)}\;    \STATE {Alternative\_epochs.\textbf{append($alt\_epoch$)}}\;
    \STATE {explore.\textbf{remove($alt\_epoch - 1$, $alt\_epoch$, $alt\_epoch + 1$)}}\;
    \FOR{$\varepsilon \in \{0, 0.01, ... , 1\}$}
    \STATE {$prob_A \gets \mathbf{mean}(class\_probs[A_i-w:A_i+w])$}\;
    \STATE {$combined\_prob \gets \varepsilon*prob_A+(1-\varepsilon)*prob$}\;
    \IF{\textbf{validation\_acc}($combined\_prob$) $\geq$ \textbf{validation\_acc}($prob$)}
    \STATE {$best\_prob = combined\_prob$}\;
    \STATE {$best\_epsilon = combined\_prob$}\;
    \ENDIF
    \ENDFOR
    \STATE {$prob = best\_prob$}\;
    \STATE {epsilons.append(\textbf{argmax}($best\_epsilon$))}\;
   \ENDWHILE
    \STATE{\textbf{Return} Alternative\_epochs, epsilons};
\end{algorithmic}
\label{alg:val}
\end{algorithm}

\begin{figure*}[htbp]
    \centering
    \begin{subfigure}[b]{0.23\linewidth}
        \includegraphics[width=\linewidth]{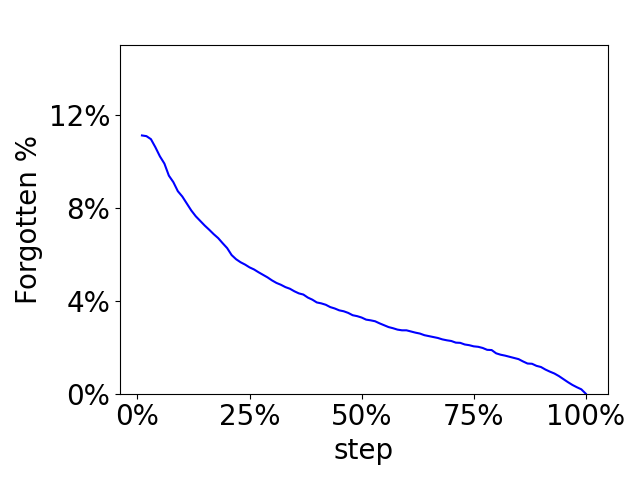}
        \includegraphics[width=\linewidth]{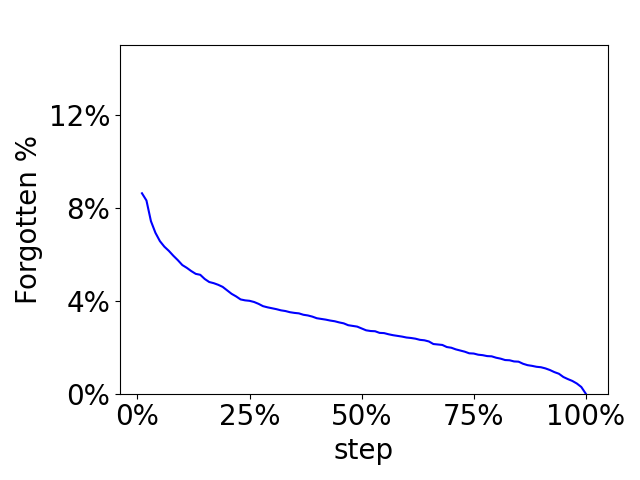}
    \vspace{-.6cm}
        \caption{$\overset{\textcolor{white}{\Sc(k)}}{\Sc(k)}$}
        \label{subfig:FRSCLRITG}
    \end{subfigure}
    \begin{subfigure}[b]{0.23\linewidth}
        \includegraphics[width=\linewidth]{figures/PC_Bias_original_ITG_forgotten_rate.png}
        \includegraphics[width=\linewidth]{figures/PC_Bias_original_ITG_forgotten_rate.png}
    \vspace{-.6cm}
        \caption{$\overset{\textcolor{white}{\Sc(k)}}{\Mc(n)}$}
    \end{subfigure}
    \begin{subfigure}[b]{0.23\linewidth}
        \includegraphics[width=\linewidth]{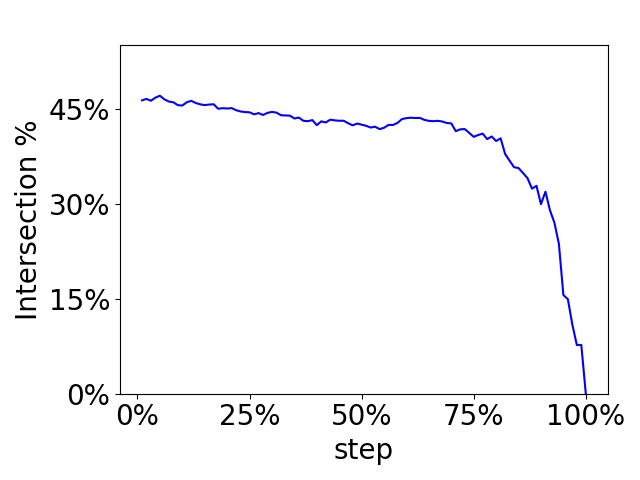}
        \includegraphics[width=\linewidth]{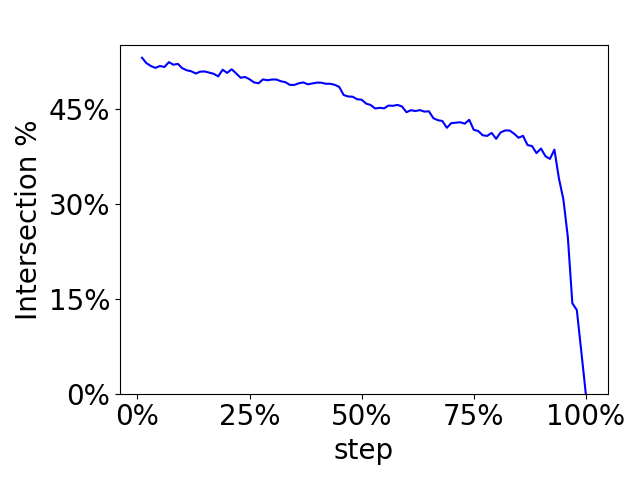}
    \vspace{-.6cm}
        \caption{\normalsize$\frac{\vert \Sc(k)\cap \Mc(n)\vert }{\vert \Sc(k)\vert}$}
        \label{subfig:IRPCSCLRITG}
    \end{subfigure}
    \begin{subfigure}[b]{0.23\linewidth}
        \includegraphics[width=\linewidth]{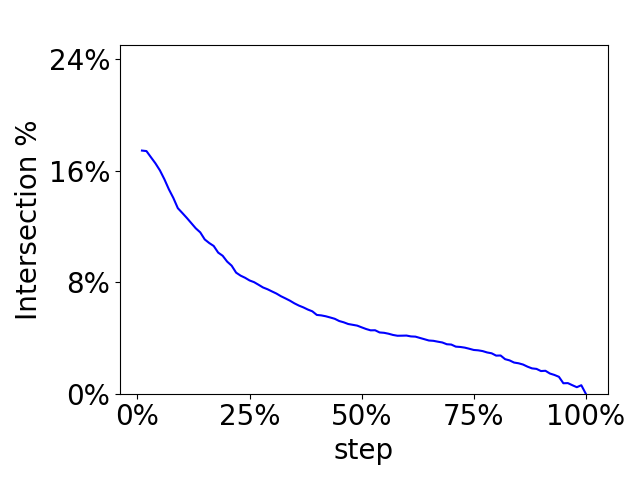}
        \includegraphics[width=\linewidth]{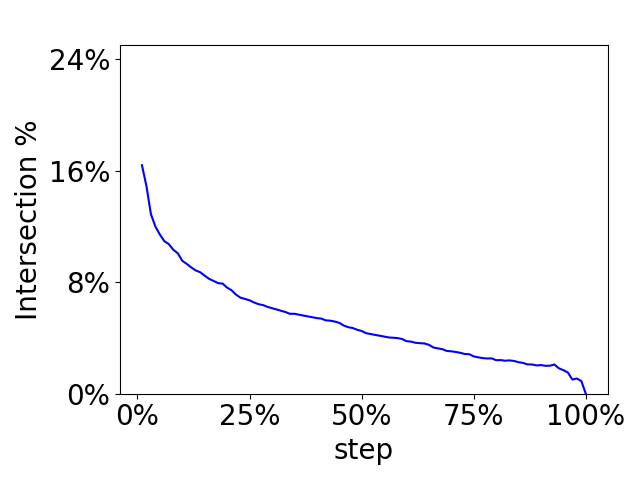}
    \vspace{-.6cm}
        \caption{\normalsize$\frac{\vert \Sc(k)\cap \Mc(n)\vert }{\vert \Mc(n)\vert}$}
        \label{subfig:IROrgSCLRITG}
    \end{subfigure}

     \caption[heatmap]{Empirical results, matching the examples forgotten during the training of a DNN and those forgotten during the training of a deep linear network. Two features spaces are used, computed with: top - the self-supervised method described in \citep{chen2020simple}; bottom - representation based on transfer learning from a ResNet50 network trained on ImageNet.}
     
     \label{fig:lastforgottenLinear}
\end{figure*}

\section{Baseline Methods Used for Comparison}
\label{app:base}

\begin{itemize}[leftmargin=0.25cm]
    \item \textbf{Snapshot ensemble} \citep{huang2017snapshot}: in this method the network is trained in several "cycles", each ending with a large increase of the learning rate that pushes the network away from the local minimum. The network is meant to converge to several different local minima during training, which are used as an ensemble. 
    \item \textbf{Stochastic Weight Averaging} (SWA) \citep{izmailov2018averaging}: in this method the network is regularly trained for a fixed training budget of epochs, and is then trained using a circular/constant learning rate to converge to several local minima, whose weights are averaged to get the final predictor. To achieve a fair comparison vis-a-vis training budget, we train the network using our training method for 75\% of the epochs, followed by their unique training for the remaining 25\% epochs. 
    \item \textbf{Fast Geometric Ensembling} (FGE) \citep{garipov2018loss}: similar to SWA, except that the final predictor is constructed by averaging the probability outputs of each model, instead of their weights. In this comparison we match budgets as explained above.
\end{itemize}

\section{Full Implementation Details}
\label{app:implementationdetails}

In our experiments we use three image classification datasets - Imagenet \citep{deng2009imagenet}, CIFAR-100 \citep{krizhevsky2009learning} and TinyImagenet \citep{le2015tiny}. In the Imagenet experiments, we train all networks (Resnet \citep{he2016deep}, ConvNeXt \citep{liu2022convnet}, ViT \citep{dosoViTskiy2020image} and MaxViT \citep{tu2022MaxViT}) using the torchvision code\footnote{https://github.com/pytorch/vision/tree/main/references/classification} for training \citep{torchvision2016} with the recommended hyper-parameters, except ConvNeXt, which was trained using the official code\footnote{https://github.com/facebookresearch/ConvNeXt} with the recommended hyper-parameters, without using exponential moving average (EMA) (see \app\ref{abl:ema} for comparison to using EMA). With CIFAR-100 and TinyImagenet, we train all networks for 200 epochs. For the clean versions of CIFAR-100 and TinyImagenet we use batch-size of 32, learning rate of 0.01, SGD optimizer with momentum of 0.9 and weight decay of 5e-4, cosine annealing scheduler, and standard augmentations (horizontal flip, random crops). 

We use similar settings for our transfer learning experiments, in which the images are resized to $224 \times 224$, the learning rate is set to 0.001 and the network is initialized using Imagenet weights. In training, either the whole network is finetuned or only a new head (instead of the original fully connected layer), which consists of two dense layers, the first with output size of 100 times the embedding size. 

For noisy labels experiments, we train using cosine annealing with warm restarts (restarting the learning rate every 40 epochs), using a larger learning rate of 0.1 and updating it after every batch. We also use a larger batch size of 64 in CIFAR-100 and 128 in TinyImagenet. In the suboptimal training described in Section~\ref{abl:subopttrain}, each image was cut before training into its central 224 over 224 pixels (images smaller than this size were first resized such that the smallest dimension was of size 224, then cut into 224 over 224). 

To obtain a fair comparison, we train the competitive methods in Table~\ref{table:specialmethods} from scratch using our network architecture and data. For \citep{huang2017snapshot} we train as instructed by the paper, while for \citep{izmailov2018averaging, garipov2018loss} we use our training scheme (as these methods are meant to be added to an existing training scheme) and performed hyper-parameter search to optimize the methods' performance in the new setting. Experiments were conducted on a cluster of GPU type A5000.

\myparagraph{Injecting label noise} For the label noise experiments we inject noisy labels using two standard methods \cite{patrini2017making}:
\begin{enumerate}
[leftmargin=0.65cm,noitemsep]
\item \textbf{Symmetric noise:} a fraction $p \in \{0.2, 0.4, 0.6\}$ of the labels is selected randomly. Each selected label is switched to any other label with equal probability. 
\item \textbf{Asymmetric noise:} a fraction $p$ of the labels is selected randomly. Each selected label is switched to another label using a deterministic permutation function. 
\end{enumerate}
\section{Additional evaluations}
\label{app:additionaleval} 
We show here additional evaluation settings of our method and baselines on dataset with injected label noise, see Table~\ref{table:additional results}.

\begin{table}[thb!]
\footnotesize
  \centering
  \begin{tabular}{l| c||c|c}
    \multicolumn{1}{ c |}{Method/\textbf{Dataset}} & \multicolumn{1}{ c ||}{\textbf{CIFAR-100 sym}} & \multicolumn{2}{ c }{\textbf{TinyImagenet}} 
    \\ 
    \multicolumn{1}{ r |}{\% label noise} &     60\% & 20\% & 40\% \\
    \toprule
        \emph{FGE}   & $38.3 \pm .7$ & $53.8 \pm .1$&$40.4 \pm .3$ \\
        \emph{SWA}   & $30.5 \pm .7$ & $52.5 \pm .2$&$39.4 \pm .3$ \\
        \emph{snapshot}   & $55.6 \pm .2$ & $62.6 \pm .1$ & $56.5 \pm .3$ \\
        \emph{KF (ours)}   & $\mathbf{57.6 \pm .2}$ & $\mathbf{62.8 \pm .2}$ &$\mathbf{57.0 \pm .5}$\\
  \end{tabular}
  \vspace{0.2cm}
  \caption{Mean (over random validation/test split) test accuracy (in percent) and standard error on image classification datasets with injected label noise, comparing our method and baselines.} 
  \label{table:additional results}
\end{table}

\section{Ablation Study}
\label{app:additionalablation}

\subsection{Using Training Set for Validation}
\label{abl:val}

In this experiment, we follow a common practice with respect to the validation data: we train our model on CIFAR-100 and TinyImagenet using only 90\% of the train data, use the remaining 10\% for validation, and finally retrain the model on the full train data while keeping the same hyper-parameters for inference. The results are almost identical to those reported in Table~\ref{table:regularnetworks}. This validates the robustness of our method to the (lack of) a separate validation set. 

\subsection{Number of Checkpoints}
\label{subsec:checkpointsnum}

\begin{figure}[htbp]
    \centering
    \begin{subfigure}[b]{0.235\linewidth}
        \includegraphics[width=\linewidth]{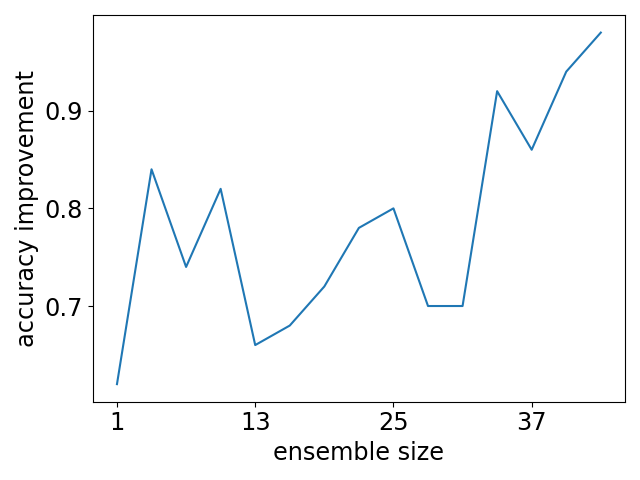}
    \vspace{-.5cm}
        \caption{CIFAR-100}
        \label{subfig:cifar100enssize}
    \end{subfigure}
    \begin{subfigure}[b]{0.235\linewidth}
        \includegraphics[width=\linewidth]{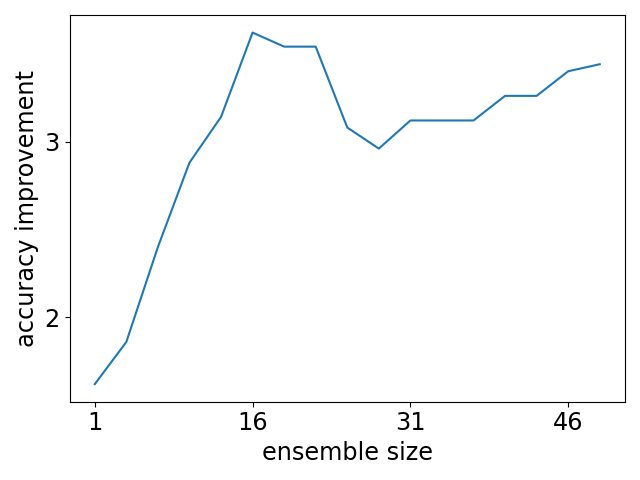}
    \vspace{-.5cm}
        \caption{TinyImagenet}
        \label{subfig:tinyenssize}
    \end{subfigure}
    \begin{subfigure}[b]{0.235\linewidth}
        \includegraphics[width=\linewidth]{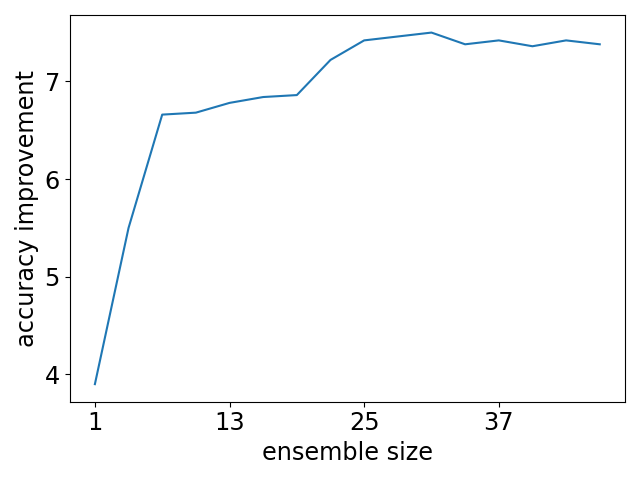}
    \vspace{-.5cm}
        \caption{20\% asym noise}
        \label{subfig:c100asym20enssize}
    \end{subfigure}
    \begin{subfigure}[b]{0.235\linewidth}
        \includegraphics[width=\linewidth]{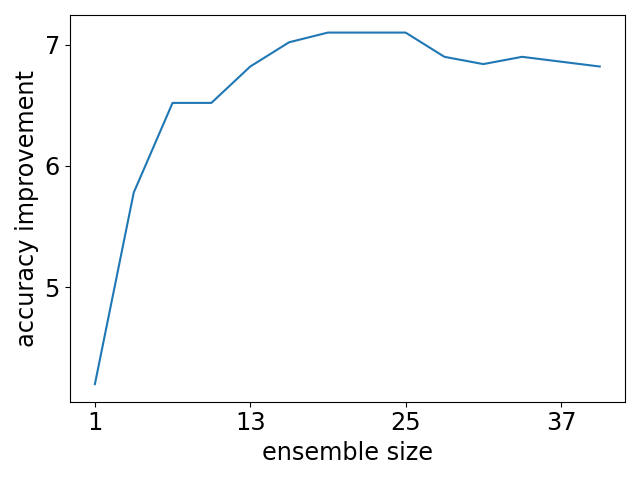}
    \vspace{-.5cm}
        \caption{20\% sym noise}
        \label{subfig:tinyenssym20size}
    \end{subfigure}
     \caption[ensemble size]{Improvement achieved by our method when using a different number of checkpoints (shown on the x-axis).
     } 
     \label{fig:ensemblesize}
         \vspace{-.5cm}
\end{figure}

Here we evaluate the cost entailed by the use of an ensemble at inference time. In Fig.~\ref{fig:ensemblesize} we report the improvement in test accuracy as compared to a single network, when varying the ensemble size. The results indicate that almost all of the improvement can be obtained using only $5-10\%$ of the checkpoints, making our method practical in real life.

\subsection{Optimal vs Sub-Optimal Training}
\label{abl:subopttrain}

In real life, full search for the optimal training scheme and hyper-parameters is not always possible, leading to sub-optimal performance. Interestingly, our method can be used to reduce the gap between optimal and sub-optimal training sessions, as seen in Table~\ref{table:suboptimal}, where this gap when training MaxViT over Imagenet is reduced by almost half when applying our method (on both models). 

\begin{table}[htbp]
\footnotesize

        \centering
  \begin{tabular}{l| c||c}
    \multicolumn{1}{ c |}{training/\textbf{method}} & \multicolumn{1}{ c ||}{\textbf{original network}} & \multicolumn{1}{ c }{\textbf{KF}} 
    \\ 
    \toprule
    \emph{regular training}   & $82.5 \pm .1$ & $83.8 \pm .1$ \\
    \emph{sub-optimal training}   & $77.3 \pm .1$ & $81.0 \pm .1$\\
    \hline
    \emph{improvement}   & $\mathbf{5.2}$ &$\mathbf{2.8}$\\

  \end{tabular}
\caption{Mean test accuracy (in percent) and ste, over random validation/test split. MaxViT is trained to classify Imagenet, comparing optimal and sub-optimal training with and without KF.} 
    \vspace{-0.25cm}

\label{table:suboptimal}

\end{table}

\subsection{Transfer Learning}
\label{abl:transferlearning}

Another popular method to improve performance and reduce overfitting employs transfer learning, in which the model weights are initialized using pre-trained weights over a different task, for example, Imagenet pretrained weights. This is followed by \emph{fine tuning} either the entire model or only some of its layers (its head for example). In Table~\ref{table:transferlearning} we show that our method is complementary to transfer learning, improving performance in this scenario as well. Note that when finetuning only the last layers, our method is \textbf{almost free of overhead costs}, as one needs to save and use at inference most of the model only once. Thus, the only overhead involves the memory and inference costs of the different head checkpoints in finetuning, whose size is insignificant as compared to the rest of the model.

\begin{table}[htbp]
\footnotesize

        \centering
  \begin{tabular}{l| c||c}
    \multicolumn{1}{ c |}{Method/\textbf{Dataset}} & \multicolumn{1}{ c ||}{\textbf{CIFAR-100}} & \multicolumn{1}{ c }{\textbf{TinyImagenet}} 
    \\ 
    \toprule
        \scriptsize{\emph{fully finetuned Resnet18} }  & $80.72 \pm .53$ & $75.01 \pm .12$ \\
        \scriptsize{\emph{fully finetuned Resnet18 + KF}}   & $\mathbf{81.64 \pm .27}$ & $\mathbf{75.6 \pm .18}$\\
        \hline
        \scriptsize{\emph{partially finetuned Resnet18}}   & $61.7 \pm .60$ & $54.78 \pm .13$ \\
        \scriptsize{\emph{partially finetuned Resnet18 + KF} }  &  $\mathbf{65.24 \pm .67}$ & $\mathbf{59.5 \pm .03}$\\
  \end{tabular}
\caption{Mean test accuracy 
     over random validation/test split.  Our method uses Resnet18 pre-trained on Imagenet, while finetuning the entire model (top) or only the head (bottom).} 

\label{table:transferlearning}

\vspace{-0.2cm}
\end{table}

\subsection{Comparisons to Additional Baselines}
\label{abl:ema}

An alternative method to combine different checkpoints is to perform exponential moving average (EMA) during training, which is known to have some advantages \citep{polyak1992acceleration} and is used sometimes to reduce overfitting \citep{liu2022convnet,dosoViTskiy2020image}, see \citep{tu2022MaxViT} for example). In Table~\ref{table:ema} we explore this option for two datasets and a regular Resnet18, showing that our method can be of use when EMA doesn't work, or improves the performance much less than our method. 

\begin{table}[thb!]
\footnotesize
  \centering
  \begin{tabular}{l| c||c}
    \multicolumn{1}{ c |}{Method/\textbf{Dataset}} & \multicolumn{1}{ c ||}{\textbf{CIFAR-100}} & \multicolumn{1}{ c }{\textbf{TinyImagenet}} 
    \\ 
    \hline
        \emph{EMA}\scriptsize{ (decay = 0.999)}   & $-0.34 \pm .14$ & $0.73 \pm .11$ \\
        \emph{EMA }\scriptsize{ (decay = 0.9999)}   & $-0.06 \pm .33$ & $2.51 \pm .01$ \\
        \emph{KF}   & $\mathbf{1.05} \pm .14$ & $\mathbf{3.54} \pm .14$\\
  \end{tabular}
  \vspace{0.2cm}
  \caption{Mean (over random validation/test split) improvement in test accuracy (in percent) and standard error on image classification datasets, comparing our method and EMA with different decay values. We use the best epoch for EMA, calculated using the validation set.} 
  \label{table:ema}
\end{table}

\begin{table*}[thb!]
\scriptsize
  \centering
  \begin{tabular}{l| c || c|c|c || c|c || c}
    \multicolumn{1}{ c |}{Method/\textbf{Dataset}} & \multicolumn{1}{ c ||}{\textbf{CIFAR-100}} & \multicolumn{3}{ c ||}{\textbf{CIFAR-100 asym}} & \multicolumn{2}{ c || }{\textbf{CIFAR-100 sym}} & \multicolumn{1}{ c }{\textbf{TinyImagenet}} 
    \\ 
    \hline
    \multicolumn{1}{ r |}{\% label noise} &    0\% & 10\% & 20\% & 40\%  &  20\% & 40\% &  0\% \\
    \toprule
    \emph{ES}   & $78.13 \pm .4$ & $71.91 \pm .2$ & $68.54 \pm .3$ & $51.53 \pm .3$ & $68.16 \pm .4$ &  $61.17 \pm .2$ & $65.44 \pm .3$ \\
    \emph{TTA}  & $79.21 \pm .3$ & $72.97 \pm .1$ & $71.00 \pm .1$ & $54.53 \pm .1$ & $70.14 \pm .2$ & $63.59 \pm .1$ & $65.67 \pm .3$ \\
    \emph{ES + TTA}  & $79.11 \pm .2$ & $73.14 \pm .2$ & $70.46 \pm .2$ & $53.93 \pm .5$ & $70.21 \pm .1$ &  $63.57 \pm .1$ & $65.74 \pm .2$ \\
    \emph{KF (ours)}  & $78.71 \pm .2$ & $73.61 \pm .1$ & $71.24 \pm .5$ & $56.19 \pm .8$ & $72.21 \pm .3$ &  $65.75 \pm .1$ & $69.00 \pm .1$ \\
    \emph{KF (ours) + TTA}  & $\mathbf{79.55  \pm .3}$ & $\mathbf{74.83 \pm .1}$ & $\mathbf{72.65 \pm .2}$ & $\mathbf{57.71 \pm .3}$ & $\mathbf{72.33 \pm .2}$ &  $\mathbf{ 65.71 \pm .1}$& $\mathbf{69.05 \pm .3}$ \\

  \end{tabular}
  \vspace{0.2cm}
  \caption{Mean test accuracy (in percent) and standard error of resnet 18, comparing our method with Early Stopping (ES) and Test Time Augmentation (TTA) on datasets with and without label noise.} 

  \label{table:tta}
\end{table*}

\begin{table*}[thb!]
\scriptsize
  \centering
  \begin{tabular}{l| c|c|c||c|c|c}
    \multicolumn{1}{ c |}{Dataset/\textbf{Method}} & \multicolumn{3}{ c ||}{\textbf{original model}} & \multicolumn{3}{ c }{\textbf{KF}} 
    \\ 
        \hline
    \multicolumn{1}{ r |}{evaluation metric} & natural accuracy&  transformed accuracy & bias & natural accuracy& transformed accuracy & bias \\
    \hline
        \emph{cifar10 w/o color}   & $89.07 \pm .48$ & $87.98 \pm .38$ & $0.07 \pm .001$ & $89.90 \pm .40$ & $87.85 \pm .48$ & $0.07 \pm .002$ \\
        \emph{cifar10 center cropped to 28x28}   & $88.45 \pm .31$ & $70.44 \pm .44$ & $0.13 \pm .003$ & $88.92 \pm .32$ & $70.21 \pm .74$ & $0.13 \pm .004$\\
        \emph{cifar10 downsampled to 16x16}   & $85.43 \pm .32$ & $76.70 \pm .13$ & $0.08  \pm .001$ & $86.55 \pm .27$ & $77.47 \pm 14$ & $0.07  \pm .001$\\
        \emph{cifar10 downsampled to 8x8}   & $80.061 \pm .33$ & $52.03 \pm .49$ & $0.22 \pm .002$ & $81.48 \pm .44$ & $52.99 \pm .49$ & $0.21 \pm .003$\\
        \emph{cifar10 with Imagenet replacements}   & $88.45 \pm .31$ & $70.44  \pm .44$ & $0.13 \pm .003$ & $88.92 \pm .32$ & $70.44 \pm .44$ & $0.13 \pm .004$\\
  \end{tabular}
  \vspace{0.2cm}
  \caption{Mean (over random validation/test split) test accuracy and amplification bias (in percent) and standard error on natural and transformed test sets, comparing our method and the original model.} 
  \label{table:fairness}
\end{table*}

Our analysis focused, for the most part, on overfitting that exists even in the scenario in which test accuracy does not decrease as training proceeds - which means that "early stopping" - culminating the training when performance over validation data decreases - has a minimal effect. Still, we wanted to compare our performance to early stopping (ES) on datasets with and without label noise. We also included in this comparison the test-time augmentation (TTA) method, in which a test example is being classified several times, each time with a different augmentation, and given a final classification based on the average class probabilities of the different classifications. The results in Table~\ref{table:tta} indicate that our method is comparable or better than both methods even when label noise exists in the training dataset (which leads to deteriorating performance as training proceeds), and that it is complementary to test-time augmentation.

\subsection{Model Size}
\label{subsec:modelsize}

\begin{table}[thb!]
\scriptsize
  \centering
  \begin{tabular}{l| c||c||c}
    \multicolumn{1}{ c |}{Method/\textbf{model size}} & \multicolumn{1}{ c ||}{\textbf{small}} & \multicolumn{1}{ c || }{\textbf{base}} & \multicolumn{1}{ c }{\textbf{large}} 
    \\ 
    \hline
        \emph{single network}   & $83.21 \pm .01$ & $83.31 \pm .15$ & $82.92 \pm .09$ \\
        \emph{KF (ours)}   & $83.17 \pm .04$ & $\mathbf{83.57 \pm .15}$ &$\mathbf{83.96 \pm .09}$\\
  \end{tabular}
  \vspace{0.2cm}
  \caption{Mean (over random validation/test split) test accuracy (in percent) and standard error on image classification datasets, comparing our method and the original predictor (ConvNeXt, trained on Imagenet) with varying number of parameters} 

  \label{table:modelsize}
\end{table}

A common practice nowadays is to use very large neural networks, with hundreds of millions parameters, or even more. However, enlarging models does not always improve performance, as a large number of parameters can lead to overfitting. In Fig.~\ref{subfig:modelsizeforget} we show that indeed larger versions of a model can cause increasing forget fraction, which also improves the benefit of our model (see Table~\ref{table:modelsize}), making it especially useful when one uses a large model.

\subsection{Comparison to a Regular Ensemble}
\label{subsec:regens}
\begin{figure}[tbp]
    \centering
    \begin{subfigure}[b]{0.45\linewidth}
        \includegraphics[width=\linewidth]{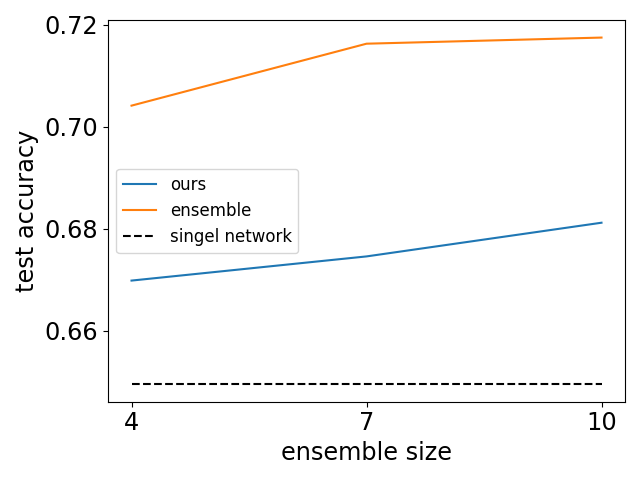}
    \vspace{-.5cm}
        \caption{TinyImagenet}
        \label{subfig:oursvsensembleTinyImg}
    \end{subfigure}
    \begin{subfigure}[b]{0.45\linewidth}
        \includegraphics[width=\linewidth]{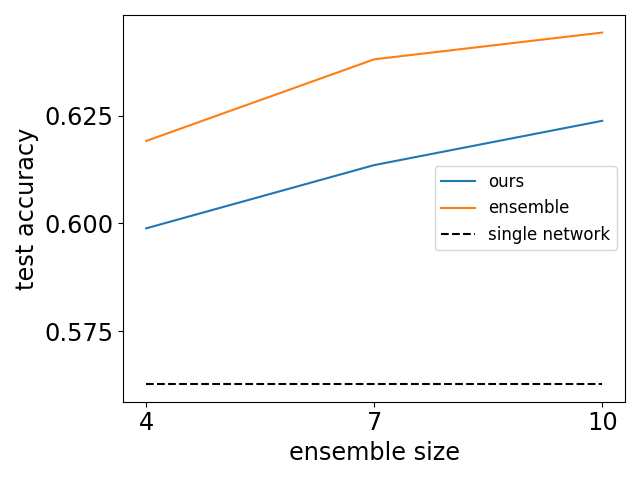}
    \vspace{-.5cm}
        \caption{TinyImagenet, 20\% sym noise}
        \label{subfig:oursvsensembleTinyImg20lb}
    \end{subfigure}
    \begin{subfigure}[b]{0.45\linewidth}
        \includegraphics[width=\linewidth]{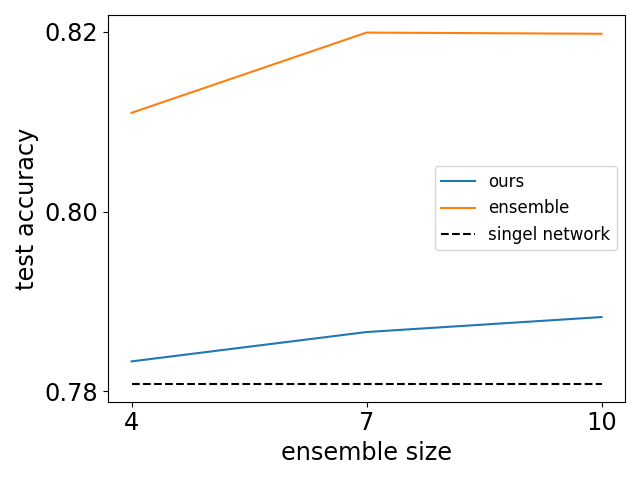}
    \vspace{-.5cm}
        \caption{CIFAR-100}
        \label{subfig:oursvsensembleC100}
    \end{subfigure}
    \begin{subfigure}[b]{0.45\linewidth}
        \includegraphics[width=\linewidth]{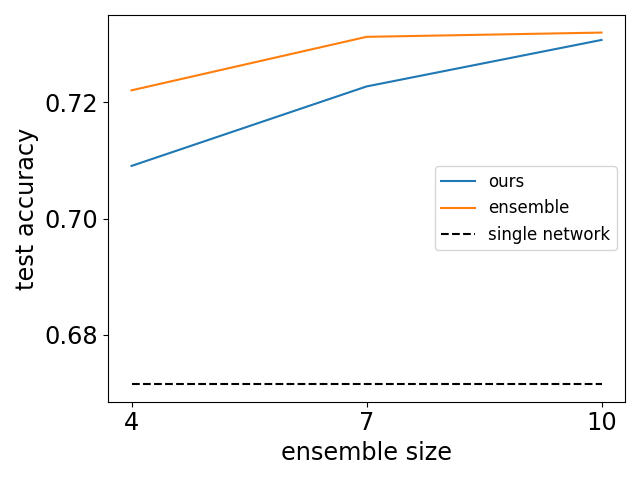}
    \vspace{-.5cm}
        \caption{CIFAR-100,  20\% asym noise}
        \label{subfig:oursvsensembleC100asym20}
    \end{subfigure}
     \caption[forgetrate]{Comparing our method with limited number of checkpoints and an ensemble of the same size of independent networks} 
     \label{fig:oursvsensemble}
\end{figure}

A regular ensemble, unlike ours, requires multiple training of independent networks, which could be unfeasible. Thus, it serves as an "upper bound" for our method's performance. In Fig.~\ref{fig:oursvsensemble}, we compare our method and a regular ensemble of the same size, showing our method can achieve much of the performance gain provided by the regular ensemble. Notably, when label noise occurs our method can add most, if not all, of the regular ensemble performance gain.
\subsection{Fairness}
\label{abl:fairness}
In this section we study our method's effect on the model's \emph{fairness}, i.e. the effect non-relevant features have on the classification of test data examples. We follow \citep{wang2020towards} and train and test our models on datasets in which they might learn spurious correlations. To create those datasets, we divide the classes into two groups: in each class of the first group 95\% of the training images goes through a transformation (and the rest remain unchanged), and vice versa for the classes of the second group. The transformation we use are: removing color, lowering the images resolution (by down sampling and up sampling), and replacing images with downsampled images for the same class in Imagenet. We use cifar10 in our evaluation as done in  \citep{wang2020towards}, and use also CIFAR-100 with the remove color transformation (the rest of the transformations were less appropriate for this datasets, as it contains similar classes that could actually become harder to seperate at a lower resolution). We use the same method as before for our validation data, and thus the validation is of the same distribution as the test data.

Our evaluation use the following metrics: (i) the test accuracy on two test sets (with/out the transformation), which should be lower if the model learns more spurious correletions, and (ii) the amplification bias defined in \citep{zhao2017men}, which is defined is follows:

\begin{equation}
    \frac{1}{|C|}\sum_{c \in C} \frac{max(c_T, c_N)}{c_T+c_N} - 0.5
\end{equation}

When $C$ is the group of classes, $c_T$ is the number of images from the transformed test set predicted to be of class c, and $c_N$ is the number of images from the natural test set predicted to be of class c - we would like those to be as close as possible, since the transformation shouldn't change the prediction, and thus the lower the score the better. 

The results of our evaluation are presented in Table~\ref{table:fairness}. To summarize, our method improves the average performance on both datasets without deteriorating the amplification bias, which indicates that our method has no negative effects on the model's fairness.

\subsection{Window Size}
\label{abl:window}
The proposed algorithm introduces a question regarding the selection of the window size ($w$) around the chosen epochs. In this work, we present results that illustrate the benefit of employing such window. Our findings demonstrate that in all evaluated scenarios, using a window greater than zero consistently outperforms the absence of a window:
\begin{figure}[htbp]
    \centering
    \begin{subfigure}[b]{0.235\linewidth}
        \includegraphics[width=\linewidth]{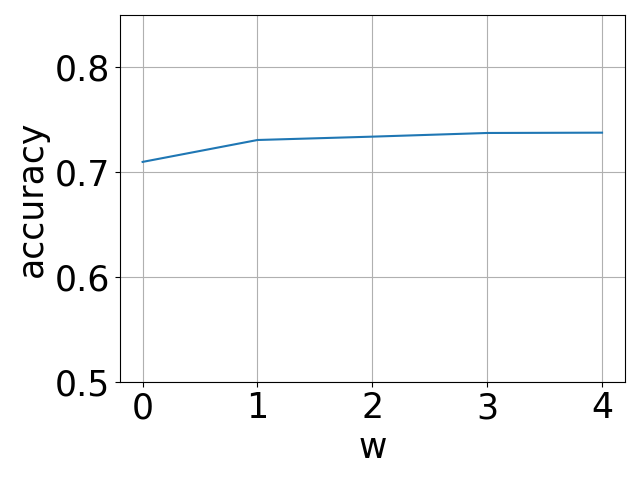}
    \vspace{-.5cm}
        \caption{20\% sym noise}
        \label{subfig:0.2symnoiseC100}
    \end{subfigure}
    \begin{subfigure}[b]{0.235\linewidth}
        \includegraphics[width=\linewidth]{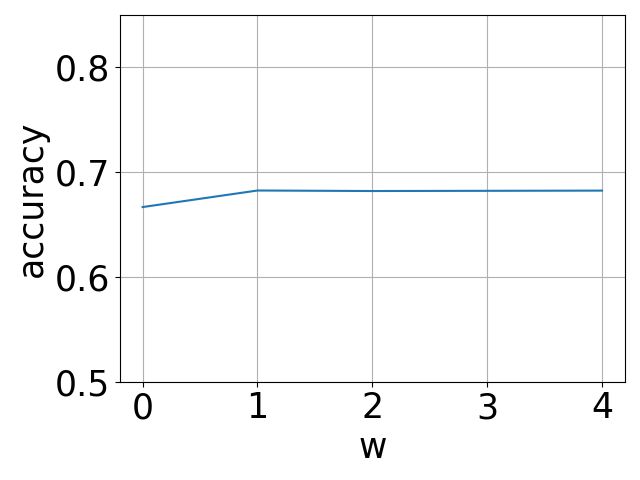}
    \vspace{-.5cm}
        \caption{40\% sym noise}
        \label{subfig:0.4symnoiseC100}
    \end{subfigure}
    \begin{subfigure}[b]{0.235\linewidth}
        \includegraphics[width=\linewidth]{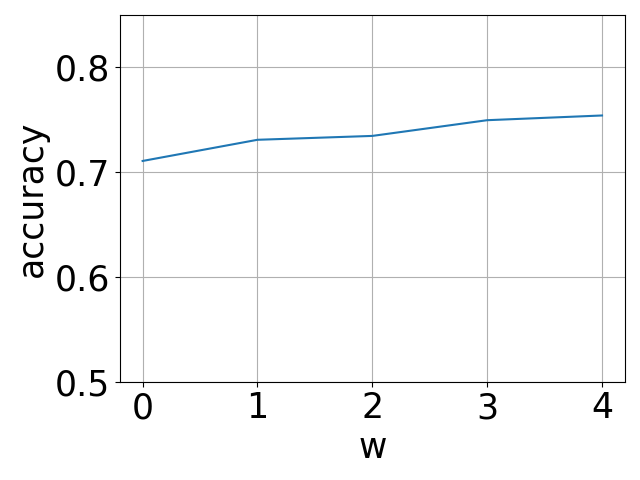}
    \vspace{-.5cm}
        \caption{20\% asym noise}
        \label{subfig:0.2asymnoiseC100}
    \end{subfigure}
    \begin{subfigure}[b]{0.235\linewidth}
        \includegraphics[width=\linewidth]{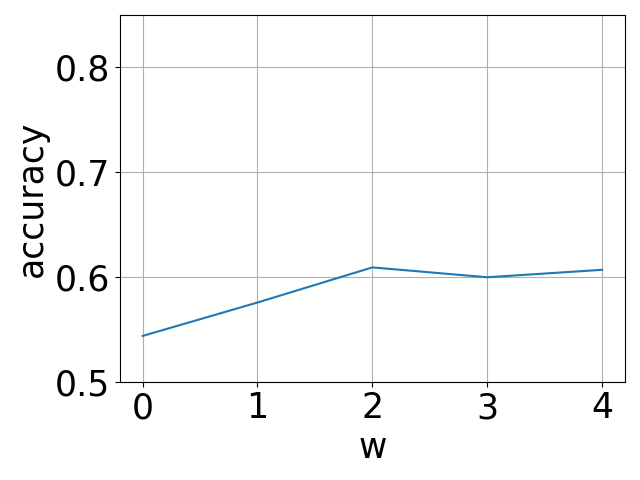}
    \vspace{-.5cm}
        \caption{40\% asym noise}
        \label{subfig:0.4asymnoiseC100}
    \end{subfigure}
     \caption[windowsize]{Accuracy of the Knowledge Fusion algorithm as a function of window size ($w$), evaluated on the CIFAR-100 dataset under varying noise levels. The results demonstrate that, irrespective of the noise level, the algorithm achieves near-optimal performance for $w\geq1$.} 
     \label{fig:windowsize}
\end{figure}

\end{document}